\documentclass[10pt,twocolumn,letterpaper]{article}

\usepackage[pagenumbers]{config/cvpr}      %

\usepackage{times}
\usepackage{epsfig}
\usepackage{graphicx}
\usepackage{amsmath}
\usepackage{amssymb}
\usepackage{caption}
\usepackage{enumitem}
\usepackage{booktabs}

\usepackage{caption}
\usepackage{makecell}
\usepackage{multirow}
\usepackage{xcolor, colortbl}
\usepackage{cuted}
\usepackage{capt-of}
\usepackage{comment}
\usepackage[normalem]{ulem} %
\usepackage{pifont}

\usepackage[pagebackref=true,breaklinks=true,colorlinks,bookmarks=false,pagebackref]{hyperref}

\usepackage[normalem]{ulem}

\usepackage{afterpage}

\newcommand{\modelname}{\mbox{\textcolor{black}{POCO}}\xspace}
\newcommand{\modelnameHPS}{\mbox{POCO-HPS}\xspace}
\newcommand{\rle}{\mbox{\textcolor{black}{RLE}}\xspace}
\newcommand{\RLE}{\rle}
\newcommand{\condnf}{\mbox{Cond-bDF}\xspace}
\newcommand{\condsc}{\mbox{Cond-Scale}\xspace}

\newcommand{\projectURL}{\href{https://poco.is.tue.mpg.de}{\tt{https://poco.is.tue.mpg.de}}}
\newcommand{\video}{\href{https://poco.is.tue.mpg.de}{\emph{video} on our website}\xspace}

\newcommand{\smpl}{\mbox{SMPL}\xspace}

\newcommand{\hps}{\mbox{HPS}\xspace}
\newcommand{\HPS}{\hps}
\newcommand{\bDF}{\mbox{bDF}\xspace}

\newcommand{\NF}{\mbox{\textcolor{black}{NF}}\xspace}
\newcommand{\DualCond}{\mbox{DCS}\xspace}
\newcommand{\DualCondFull}{Dual Conditioning Strategy\xspace}

\newcommand{\hypothMulti}{\textcolor{black}{multiple        bodies}\xspace}

\newcommand{\captionUnits}{All metrics are in mm}

\newcommand{\cheading}[1]{\noindent\textbf{#1.}}
\newcommand{\qheading}[1]{\noindent\textbf{#1:}}

\newcommand{\zheading}[1]{\textbf{#1:}}

\usepackage{lipsum}
\usepackage{balance}

\newcommand{\BLdeterministicHPS}{{Baseline \HPS}\xspace}
\newcommand{\BLGauss}{{Gauss}\xspace}
\newcommand{\BLNFlow}{{NFlow}\xspace}

\definecolor{citecolor}{HTML}{0071bc}
\definecolor{frontcolor}{HTML}{325ea5}
\definecolor{sidecolor}{HTML}{a58b77}
\definecolor{DeltaColor}{rgb}{0.039,0.73,0.71}
\definecolor{SigmaColor}{rgb}{0.98,0.45,0.0}
\definecolor{AlphaColor}{rgb}{0,0,0.8}
\definecolor{BetaColor}{rgb}{0.8,0,0.8}
\definecolor{GammaColor}{rgb}{0.514,0.34,0.224}
\definecolor{EpsilonColor}{rgb}{0.353,0.725,0.906}
\definecolor{PurpleColor}{HTML}{bca5ea}
\definecolor{OrangeColor}{rgb}{0.914,0.541,0.0.141}
\definecolor{GreenColor}{rgb}{0.137,0.573,0.565}
\definecolor{RedColor}{rgb}{0.949,0.275, 0.224}
\definecolor{LightCyan}{rgb}{0.88,1,1}
\definecolor{Gray}{gray}{0.85}

\newcommand{\confidence}{\textcolor{black}{confidence}\xspace}

\newcommand{\uncertainties}{uncertainties\xspace}
\newcommand{\uncertainty}{uncertainty\xspace}
\newcommand{\Uncertainty}{Uncertainty\xspace}
\newcommand{\uncertain}{uncertain\xspace}

\newcommand{\dataColor}{black}
\newcommand{\threeDOH}{\mbox{\textcolor{\dataColor}{3DOH}}\xspace}
\newcommand{\threedpw}{\mbox{\textcolor{\dataColor}{3DPW}}\xspace}
\newcommand{\threeDPW}{\threedpw}
\newcommand{\threeDPWOCC}{\mbox{\textcolor{\dataColor}{3DPW-OCC}}\xspace}
\newcommand{\mpiinf}{\mbox{MPI-INF-3D}\xspace}
\newcommand{\coco}{\mbox{COCO}\xspace}
\newcommand{\mpii}{\mbox{MPII}\xspace}
\newcommand{\lspet}{\mbox{LSPET}\xspace}
\newcommand{\humanThreeSixFull}{\mbox{Human3.6M}\xspace}
\newcommand{\humanthreesix}{\mbox{H3.6M}\xspace}

\newcolumntype{a}{>{\columncolor{Gray}}c}

\newcommand{\moveToSupMat}[1]{\begin{comment}#1\end{commment}}
\newcommand{\supmat}{\textcolor{magenta}{\emph{Sup.~Mat.}}\xspace}

\newcommand{\resnet}{\mbox{ResNet}\xspace} %

\newcommand{\deterM}{\mbox{Dtr}\xspace}
\newcommand{\probM}{\mbox{Prob}\xspace}

\newcommand{\xmark}{\textcolor{RedColor}{\ding{55}}\xspace}
\newcommand{\cmark}{\textcolor{GreenColor}{\ding{51}}\xspace}

\newcommand{\colorRef}[1]{\textcolor{red}{#1}} %
\usepackage[capitalize]{cleveref}
\crefname{figure}{\colorRef{Fig.}}{\colorRef{Figs.}}
\Crefname{figure}{\colorRef{Figure}}{\colorRef{Figures}}
\crefname{section}{\colorRef{Sec.}}{\colorRef{Secs.}}
\Crefname{section}{\colorRef{Section}}{\colorRef{Sections}}
\Crefname{table}{\colorRef{Table}}{\colorRef{Tables}}
\crefname{table}{\colorRef{Tab.}}{\colorRef{Tabs.}}
\Crefname{equation}{\colorRef{Equation}}{\colorRef{Equations}}
\crefname{equation}{\colorRef{Eq.}}{\colorRef{Eqs.}}

\renewcommand{\etal}{\mbox{et al.}\xspace}
\renewcommand{\ie}{\mbox{i.e.}\xspace}
\renewcommand{\eg}{\mbox{e.g.}\xspace}

\newcommand{\groundtruth}{{ground-truth}\xspace}

\newcommand{\pseudoGT}{\mbox{pseudo-GT}\xspace}

\newcommand{\inthewild}{\mbox{in-the-wild}\xspace}

\newcommand{\ood}{\mbox{out-of-distribution}\xspace}

\newcommand{\twoD}{2D\xspace}
\newcommand{\threeD}{3D\xspace}
\newcommand{\fourD}{4D\xspace}

\newcommand{\hmr}{\mbox{HMR}\xspace}
\newcommand{\hmreft}{\mbox{HMR-EFT}\xspace}
\newcommand{\eft}{\mbox{EFT}\xspace}
\newcommand{\pare}{\mbox{PARE}\xspace}
\newcommand{\PARE}{\pare}
\newcommand{\glamr}{\mbox{GLAMR}\xspace}

\newcommand{\cliffOur}{\mbox{CLIFF-Ours}\xspace}
\newcommand{\cliff}{\mbox{CLIFF}\xspace}

\newcommand{\regPose}{\theta}
\newcommand{\gtPose}{\theta_g}
\newcommand{\bDFPose}{\bar{\theta}}
\newcommand{\nfPose}{\hat{\theta}}
\newcommand{\gtDevPose}{\bar{\theta}_g}

\newcommand{\regScale}{\sigma}

\newcommand{\regShape}{\beta}
\newcommand{\gtShape}{\beta_g}

\newcommand{\cam}{C}

\newcommand{\threeDJnts}{J_{3D}}
\newcommand{\twoDJnts}{J_{2D}}
\newcommand{\gtThreeDJnts}{J_{3D_g}}
\newcommand{\gtTwoDJnts}{J_{2D_g}}
\newcommand{\JntReg}{J_{Reg}}
\newcommand{\mesh}{\mathcal{M}}
\newcommand{\vertices}{V}

\newcommand{\image}{I}
\newcommand{\imageFeats}{I_{\mathit{feats}}}
\newcommand{\InputcnfNN}{C_{\mathit{nf}}}

\newcommand{\regressor}{\Theta}
\newcommand{\CNF}{\Phi}
\newcommand{\cnfNN}{\mathit{NN}_{\mathit{nf}}}
\newcommand{\cscNN}{\mathit{NN}_{\mathit{sc}}}

\def\paperID{44} %
\def\confName{3DV\xspace}
\def\confYear{2024\xspace}

\begin{document}

\title{\vspace{-0.5 em}\modelname: \threeD Pose and Shape Estimation with Confidence\vspace{-0.3 em}}

\author{
    Sai Kumar Dwivedi$^{1}$\quad \; 
    Cordelia Schmid$^{2}$\quad \; 
    Hongwei Yi$^{1}$\quad \; 
    Michael J. Black$^{1}$\quad \; 
    Dimitrios Tzionas$^{3}$\\
    \normalsize $^1$Max Planck Institute for Intelligent Systems, T\"{u}bingen, Germany \\ 
    \normalsize $^2$Inria, \'Ecole normale sup\'erieure, CNRS, PSL Research University, France\\
    \normalsize $^3$University of Amsterdam, the Netherlands\\
}

\maketitle

\begin{strip}
    \centering
 	\vspace{-2.5 em}
 	\includegraphics[width=\textwidth]{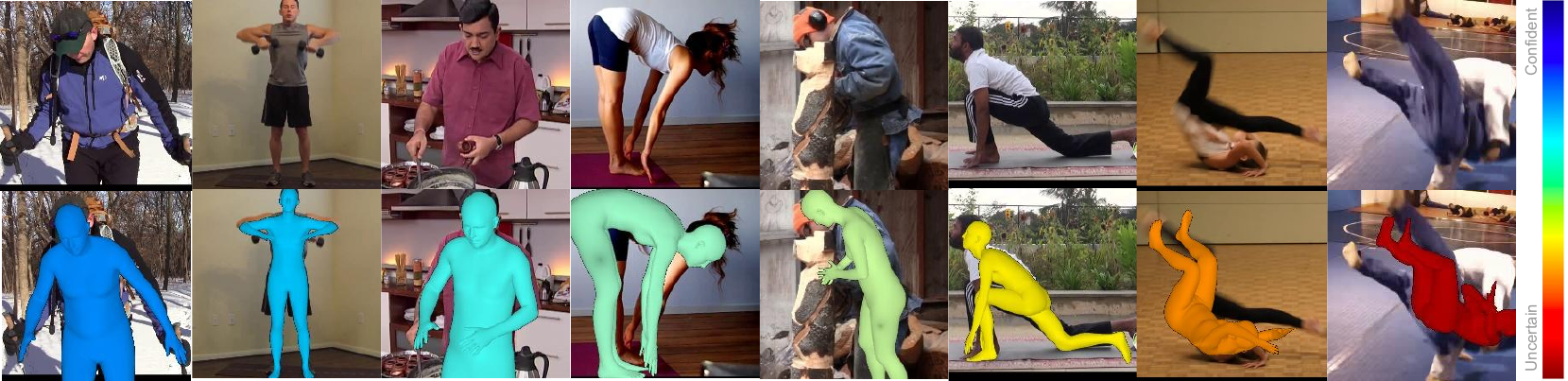}
 	\vspace{-2.0em}
 	\captionsetup{type=figure}
  	\captionof{figure}{
  	    \cheading{Regressing \threeD humans and \confidence}
          Most existing methods that regress 3D Human Pose and Shape (\HPS) do not report their \confidence (or \uncertainty).
          An estimate of confidence, however, is needed by methods that ``consume" the results of HPS.
          Even the best HPS methods struggle when the image evidence is weak or ambiguous. %
  	 Our framework, \modelname, extends existing \hps regressors to also estimate \uncertainty in a single forward pass.
  	 The \confidence (color-coded on the body) is correlated with 
          the pose quality.
    }
 	\label{fig:teaser}
 	\vspace{+1.0em}
\end{strip}

\begin{abstract}
\vspace{-1.0em}
The regression of \threeD Human Pose and Shape (\HPS) from an image is becoming increasingly accurate.
This makes the results useful for downstream tasks like human action recognition or 3D graphics.
Yet, no regressor is perfect, and accuracy can be affected by ambiguous image evidence or by poses and appearance that are unseen during training.
Most current \HPS regressors, however, do not report the confidence of their outputs, meaning that downstream tasks cannot differentiate accurate estimates from inaccurate ones.
To address this, we develop \textit{\modelname}, a novel framework for training \hps regressors to estimate not only a \threeD human body, but also their \confidence, in a single feed-forward pass.
Specifically, \modelname estimates both the \threeD body pose and a per-sample variance.
The key idea is to introduce a \textit{\DualCondFull} (\DualCond) 
for regressing 
\uncertainty that is highly correlated to pose reconstruction quality.
The \modelname framework can be applied to any \hps regressor and here we evaluate it by modifying HMR, PARE, and CLIFF.
In all cases, training the network to reason about \uncertainty helps it learn to more accurately estimate 3D pose.
While this was not our goal, the improvement is modest but consistent. 
Our main motivation is to provide \uncertainty estimates for downstream tasks; 
we demonstrate this in two ways:  
(1) We use the confidence estimates to bootstrap \hps training.
Given unlabeled image data, we take the confident estimates of a \modelname-trained regressor as pseudo ground truth.
Retraining with this automatically-curated data improves accuracy. 
(2) We exploit \uncertainty in video pose estimation by automatically identifying uncertain frames (e.g.~due to occlusion) and ``inpainting" these from confident frames.
Code and models will be available for research at 
\projectURL.
\vspace{-1.5 em}
\end{abstract}

\section{Introduction}

To reconstruct human actions in everyday environments we need to estimate \threeD Human Pose and Shape (\HPS) from images and videos. 
However, \threeD inference from \twoD images is highly ill-posed due to depth ambiguities, occlusion, unusual clothing, and motion blur.
Not surprisingly, even the best \HPS methods make mistakes.
The problem is that they do not know it and neither do downstream tasks.
\HPS is not an end goal but, rather, an intermediate task that produces output that is consumed by downstream tasks like human behavior understanding or \threeD graphics applications.
Downstream tasks need to know when the results of an \HPS method are accurate or not.
Consequently, these methods should output a \uncertainty (or \confidence) value that is \emph{correlated with the \HPS quality}. 

\pagebreak

One approach to dealing with \uncertainty is to output \hypothMulti \cite{multibodies}, but this still does not provide an explicit measure of \uncertainty.
There are several notable exceptions that estimate a distribution over body parameters \cite{prohmr,hierprobhuman,li_multiskeleton};
sampling from this distribution generates multiple plausible bodies. 
For example, Sengupta \etal~\cite{hierprobhuman} compute \uncertainty by drawing samples from a distribution over bodies and computing the standard deviation of the samples.
While a valid approach, it has two drawbacks:
(1) it is slow, since it requires multiple forward network passes to generate samples, and 
(2) it trades off accuracy for speed; the more the samples, the higher the accuracy, but the more computation is required. 
Instead, we propose an approach that infers \uncertainty directly in a \emph{single network pass} by training the network to output both the body parameters and the \uncertainty.
Note that we achieve this without any explicit supervision of the \uncertainty.

We draw inspiration from Kendal \etal~\cite{kendall_uncert} who estimate uncertainty for semantic segmentation; 
they model the error of inferred segmentations
using a \textit{base density function} (\bDF); they use a Gaussian distribution for this. Then, they infer a per-sample scale (\ie, \uncertainty) using a \textit{scale network} to refine this function and model a per-sample likelihood.

A Gaussian distribution is common for modeling sample error. 
However, recent work \cite{rle,hierprobhuman,humaniflow,Zanfir:ECCV:2020} shows that a more complex distribution like a normalizing flow~\cite{realnvp} is needed when modeling human poses, \eg, \RLE \cite{rle} does so for modelling the error of 
\threeD body joints.

Methods that directly estimate \uncertainty~\cite{rle, kendall_uncert} usually have two components - a \textit{base density function} and a \textit{scale network} to refine the \bDF.
Kendal \etal~\cite{kendall_uncert} and RLE~\cite{rle} use an unconditional \bDF and only use image-features for the \textit{scale network}. 
An unconditional \bDF, shared among all samples, is reasonable if all samples share a similar distribution.
However, this assumption breaks when using multiple datasets, which are needed to train robust \threeD \HPS models.
Similarly, using only image features to infer a per-sample scale (\ie, \uncertainty) does not consider the \textit{pose plausibility} for the given image evidence.

We address these issues with \textbf{\modelname} (\textit{``\textbf{PO}se and shape estimation with \textbf{CO}nfidence''}), a novel framework that can be applied to common
\HPS methods, extending them to estimate \uncertainty.
In a single feed-forward pass, \modelname directly infers both \smpl \cite{smpl} body parameters and its regression \uncertainty~which is highly correlated to reconstruction quality; see \cref{fig:teaser}. 
The key novelty of our framework is a \textit{\DualCondFull} (\DualCond) which augments the \bDF and \textit{scale network} as described below:

\qheading{(1)~Image-conditioned \bDF}
In contrast to 
prior work,
\modelname models the \bDF of the inferred pose error with a \emph{conditional} 
vector
(\condnf). 
A naive condition would be a one-hot encoding of the data source. 
However, rich datasets contain images that are not necessarily drawn from a single distribution. 
Therefore, we use image features as conditioning, making training more scalable and possible on arbitrarily complex collections of images.

\qheading{(2)~Pose-conditioned scale}
Previous methods infer a scale (\ie, \uncertainty) value that refines the \bDF for each image, using an MLP that takes image features as input.
We go further to also condition the MLP on the regressed \smpl pose (``\condsc'').
We experimentally show that this improves pose reconstruction and helps estimate \uncertainty based on pose plausibility for given image evidence.

Our formulation can be included in the loss function of existing \HPS regressors.
We show this by modifying HMR \cite{hmr}, PARE \cite{pare}, and CLIFF \cite{cliff} to also output \uncertainty. 
A key observation is that this change uniformly improves the accuracy of these methods. 
That is, requiring the network to also estimate \uncertainty helps it learn to better estimate pose and shape.
While the accuracy improvement is modest, the important point is that one can apply \modelname to existing regressors with no downside; \ie,~one gets \uncertainty estimates ``for free."

We perform qualitative evaluation on in-the-wild data and quantitative evaluation on \threeDPW \cite{vonMarcard2018} and \threeDOH \cite{zhang3DOH}. 
We train all methods (HMR, PARE, CLIFF) and their \modelname versions on the same data and show a quantitative improvement in accuracy in all cases.
We also compare these \modelname formulations to state-of-the-art (SOTA) \HPS methods that model uncertainty and show higher correlation of uncertainty and pose error.
Ablation studies show the efficacy of \modelname's novel contributions.

While improved pose accuracy is a welcome byproduct, our main goal is to exploit \uncertainty estimates in downstream tasks.
We demonstrate this in two practical tasks. 

\vspace{-0.1 em}
\qheading{(1)~\HPS self-improvement}
We show that \HPS models can self-improve by leveraging \uncertainty estimates to automatically bootstrap training. 
Specifically, we apply an \HPS regressor trained with \modelname on in-the-wild videos for which there is no ground truth. 
Normally such data would not be useful for training. 
We then take pairs of images and \smpl parameters that have high confidence and treat them as pseudo ground-truth training samples.
We finally re-train the regressor on an enriched dataset that includes the pseudo ground truth and show that this improves accuracy. 
This works with all three \HPS regressors that we evaluated.

\vspace{-0.1 em}
\qheading{(2)~\HPS from video}
In complex videos, single-frame \HPS regressors make errors on challenging frames.
We exploit the estimated \uncertainty to automatically detect frames where the \smpl estimates may be inaccurate. 
We then remove these \uncertain estimates and infill the corresponding frames following GLAMR \cite{yuan2021glamr}.
This approach uses highly-confident bodies and knowledge of human motion to produce more plausible \fourD \HPS results.

\pagebreak

In summary, we make the following key contributions:
(1) We are the first to demonstrate a general \uncertainty framework, \modelname, that can be applied to common \hps methods for estimating \uncertainty in a single \textit{forward pass}.
(2) We introduce a \textit{\DualCondFull} (\DualCond) that helps regress \uncertainty that is highly correlated to pose reconstruction quality and improves the accuracy of exisiting \HPS regressors.
(3) We show that the \uncertainty estimate given by \modelname can be used for two practical downstream tasks.
Our framework gives existing methods a simple way to estimate \uncertainty with no downside, and it even improves performance.
We believe this will make \HPS results more useful for downstream tasks.
Our models and code will be available.
\section{Related Work}

\subsection{Deterministic \HPS Estimation}

\zheading{Optimization-based methods} 
Such methods fit a parametric model \cite{scape,smpl,smplx,xu2020ghum,Joo2018_adam} to cues extracted from images, such as keypoints \cite{smplify,Xiang_2019_CVPR,xu2020ghum}, silhouettes \cite{nbf,huang2017towards},  part segmentation masks \cite{unitepeople}, or part orientation fields \cite{Xiang_2019_CVPR}.

\zheading{Learning-based methods} 
Datasets with \twoD or \threeD ground truth have enabled
deep learning methods to regress parametric body models \cite{smpl,xu2020ghum,STAR:ECCV:2020}
from images \cite{hmr, pavlakos2018, denserac, spec, vibe, hier_mesh_recovery, dsr, eft, pare, thunder, hybrik, cliff, Sun:CVPR:2022} or videos \cite{vibe, humanMotionKanazawa19}.
Some recent methods \cite{cmr,l2l-meshnet,pose2mesh,pifu,pifuhd,icon,meshgraphormer} estimate bodies in a model-free fashion, instead of using a parametric body model.
Specifically, they either directly predict mesh vertices  \cite{meshgraphormer,cmr,l2l-meshnet,pose2mesh} or implicit surfaces \cite{pifu,icon,Mihajlovic_CVPR_2022}. 

\zheading{Hybrid methods}
To get the best of both worlds, some methods combine regression and optimization. 
A regressor provides a rough \smpl \cite{smpl} body estimate for an image, and then an optimizer refines the \smpl parameters, so that the refined body better fits \twoD joint annotations \cite{spin}. 
This can be extended so that, instead of \smpl parameters, the optimizer refines the regressor's network weights (EFT \cite{eft}) for each image separately, and the regressed body better fits \twoD joint annotations.
EFT is used to recover good pseudo ground-truth bodies from unlabeled images, while a human annotator finally curates the fits. 
The resulting data help train better \HPS regressors; 
we train \modelname on this.

Note that all of the above methods predict only \threeD bodies \emph{without} any measure of 
\uncertainty.

\subsection{Probabilistic \HPS Regression}

Early work on estimating \threeD human pose addresses \uncertainty 
by using sampling-based methods to infer human pose from images and videos \cite{sampling_old1,sampling_old2}.
Learning-based approaches predict multiple \threeD poses given \twoD cues. 
Li \etal~\cite{li_multiskeleton} infer a distribution of \threeD joints using a Mixture Density Network (MDN) conditioned on \twoD joints. 
Others use a Normalizing Flows (\NF), instead, to infer the distribution of single- \cite{multi_joints_nflow} or multi-person \cite{mp_distribution} \threeD joints. 
ProHMR \cite{prohmr} uses a \NF conditioned on image features to infer \smpl bodies. 
Biggs \etal~\cite{multibodies} infer a %
set of 
$M$ \smpl bodies, and 
train with a ``best-of-$M$'' loss. %
Sengupta \etal \cite{hierprobhuman} use 
a hierarchical matrix-Fisher distribution
over \smpl parameters.
Sengupta \etal~compute \uncertainty by sampling many bodies and computing the standard deviation. 
This makes computing \uncertainty a post processing that trades-off accuracy (\# of samples) for speed (\# of network passes). 
Instead, the \modelname framework \emph{directly} infers \uncertainty in a \emph{single network pass}.

\subsection{\Uncertainty Modeling}

Work on \uncertainty modeling falls into four categories.
{\bf Bayesian methods} 
 \cite{bayesian1,bayesian2,bayesian3} model network weights as a random variable. 
This enables sampling new network weights during the feed-forward pass. 
{\bf Ensemble methods}
 \cite{ensemble1,ensemble2} combine predictions from multiple models that are trained differently or use different input modalities, \eg, Lidar scans or images.
{\bf Test-time augmentation methods} 
\cite{testtime_un1,testtime_un2,testtime_un3} apply several data-augmentation techniques to the input and perform a prediction for each of these.
{\bf Direct inference methods} 
 \cite{deterministic_un1, deterministic_un2, deterministic_un3, deterministic_un4, kendall_uncert, rle} infer a single deterministic output and an \uncertainty value that models the output's deviation from the ground truth.

The first three categories get multiple outputs and analyze their variance to approximate \uncertainty. 
However, this trades-off accuracy for speed; the more samples drawn, the more passes are needed (making the runtime slower), but the more accurate the \uncertainty computation gets. 
Direct inference methods do not suffer from these limitations.

For semantic segmentation, Kendal \etal~\cite{kendall_uncert} define two types of \uncertainty:
\emph{aleatoric}, 
caused by ambiguities in images, and 
\emph{epistemic}, 
caused by insufficient data, which models the sample error. 
For the latter, they use a Bayesian network. 
For the former, they use a Gaussian distribution, but instead of a fixed variance, they infer a per-sample variance as an \uncertainty metric.

Direct estimation of \uncertainty for pose estimation has primarily focused on human skeletons. To estimate \threeD joints in a sequence, Zhang \etal~\cite{uncert_joints_video} use two separate Gaussian distributions; one for \twoD keypoints and one for depth. 
Kundu \etal~\cite{multijoints_ss_cvpr22} model the consistency of different pose representations as a proxy for uncertainty.
To infer human joints, \RLE~\cite{rle} assumes that all images are drawn from a single  distribution, and uses a Normalizing Flow (\NF) 
as a base density function (bDF), shared across all samples, to model sample error. 
Then, they infer a per-sample translation and scale to refine this bDF; the scale is treated as an \uncertainty metric. 
However, training robust \HPS models requires multiple datasets, which breaks the assumption that samples follow a single distribution.

\begin{figure*}
    \vspace{-0.5 em}
    \centerline{\includegraphics[width=\linewidth]{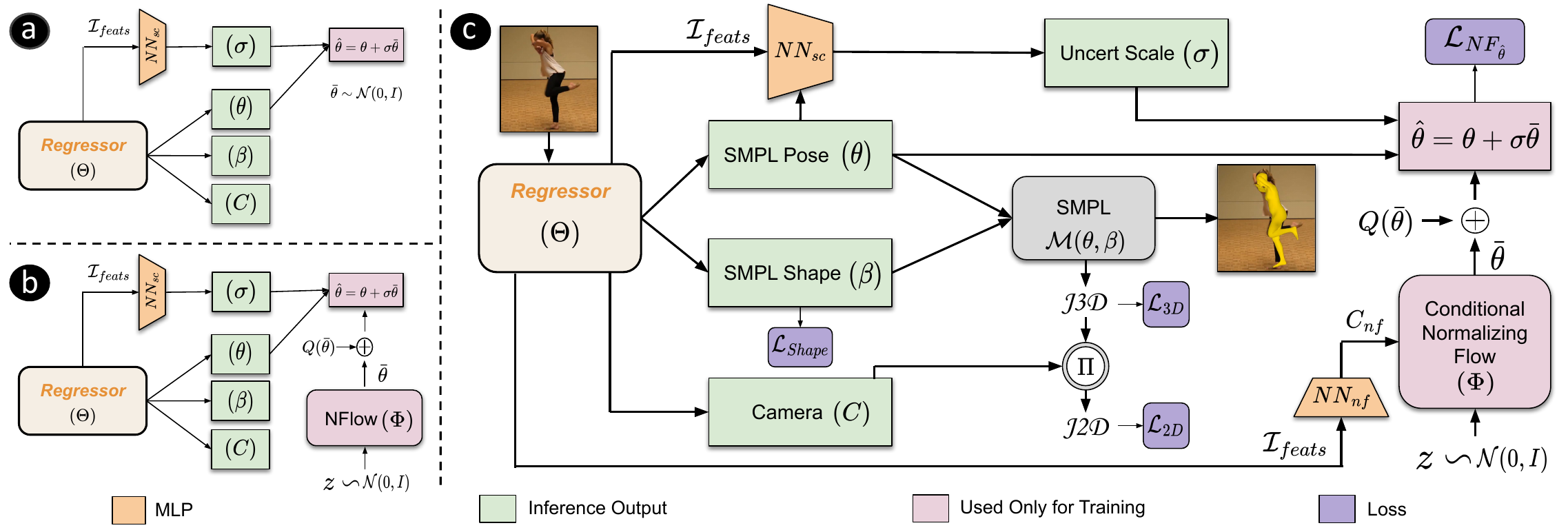}}
    \vspace{-0.8 em}
    \caption{
                \cheading{\modelname framework}
                Given an \HPS regressor, we show how to upgrade it to also infer $\sigma$, representing the output uncertainty.
                This works for most common regressors.
                \textbf{(a)} 
                This baseline model infers $\sigma$ with a Gaussian base density function (\bDF).
                \textbf{(b)} 
                This baseline replaces the Gaussian \bDF with a Normalizing Flow (\NF) shared across all samples.
                \textbf{(c)} 
                \modelname models an expressive \emph{conditional \bDF} (\condnf) and a \emph{pose-dependent} $\sigma$ (\condsc)
                that is better correlated with pose reconstruction quality.
            }
    \label{fig:architecture}
    \vspace{-1.0 em}
\end{figure*}

\modelname extends \RLE in two ways: 
(1) it conditions the \NF on image features to model an expressive per-sample bDF, and 
(2) it infers \smpl bodies instead of joints, and conditions scale inference (to refine each bDF) on \smpl pose. 
As a result, \modelname uniquely enables \emph{both} using multiple datasets and direct inference of \uncertainty for \threeD \hps.

\section{Method}

The \modelname framework starts with a base \HPS regressor and augments the network and training so that the method  takes an input image, $\image$, and outputs an \uncertainty estimate, $\regScale$, in addition to the standard \smpl pose, $\regPose$, shape, $\regShape$, and weak-perspective camera, $\cam$, parameters;
see \cref{fig:architecture} \textcolor{red}{(c)}.

\subsection{Preliminaries}

\zheading{Body model} 
\smpl \cite{smpl} is a differentiable parametric human body model. 
As input, it takes parameters for pose,  $\regPose  \in \mathbb{R}^{72}$, and shape, $\regShape \in \mathbb{R}^{10}$. 
As output, it produces a body mesh, $\mesh$, and vertices, $\vertices \in \mathbb{R}^{NX3}$, where $N = 6890$ is the number of vertices. 
\threeD skeletal joints, $\threeDJnts$, are computed from a linear combination of mesh vertices, $\vertices$, using a pre-trained linear regressor $\JntReg$, \ie,  $\threeDJnts = \JntReg\vertices$.

\zheading{Regressor for \smpl bodies} 
The core idea of  \modelname can be employed as a component of any regressor network, $\regressor(\image)$, that estimates a \smpl body and a camera from an image. 
Here we evaluate \modelname using three different regressors; \ie, \hmr \cite{hmr}, \pare \cite{pare}, and CLIFF~\cite{cliff}. 
The main difference between these is the way they compute features from the image.
\hmr uses a \resnet to compute a single global feature from the image; this often causes problems for poses and occlusions that are unseen during training. 
\pare accounts for this by using human-part-guided attention to compute several per-part features that are aggregated; this gives robustness to occlusions, as the non-occluded parts help resolve ambiguities caused by the occluded ones. 
\hmr and \pare operate on an image cropped around a body. 
In contrast, CLIFF also considers the body's location in the full image; this global context helps improve poses. 
CLIFF represents the current SOTA.
For details of each model's architecture, see \supmat

\zheading{Normalizing Flow (\NF)}
These models are used to model arbitrarily complex distributions through a composition of smooth and invertible transformations of a simple distribution. 
Let $Z \in \mathbb{R}^d$ be a random variable from a simple distribution, \eg, a multivariate Gaussian distribution, $P_Z(z)$. 
Let also $f: \mathbb{R}^d \rightarrow \mathbb{R}^d$ be a  smooth and invertible function. 
The NF transforms $z$ into a complex distribution $x = f(z)$ and, as it is invertible, we can define $z=f^{-1}(x)$. 
The log probability density function of $x$ is: 
\vspace{-0.5 em}
\begin{equation}
    \textnormal{log}P_{X}(x) = \textnormal{log}P_{Z}(z) + \textnormal{log}\left|\textnormal{det}\frac{\partial f^{-1}}{\partial z}\right|
    \text{.}
    \label{eqn:1}
\end{equation}

\subsection{Unconditional \Uncertainty Estimation} \label{sec:uncertain}

\zheading{\mbox{Sec. 3.2.1} MSE loss \& error distribution}
A standard mean-squared-error (MSE) loss, typically used for training \HPS models \cite{pare,spin}, assumes the inferred-sample error to be Gaussian distributed with a constant variance across all samples. 
Below, we refer to this Gaussian as a \emph{base density function (\bDF)} that is shared among all samples.
Kendall \etal~\cite{kendall_uncert} replace the constant variance with a per-sample variance, in the context of semantic segmentation.

We adapt the work of Kendall \etal~for \HPS as shown in ~\cref{fig:architecture}~\textcolor{red}{(a)}, and use a regressor, $\regressor$, to estimate per-sample parameters for pose, $\regPose_i$, and variance, $\regScale_i$, from an image, $\image_i$. 
We define the pose loss over a dataset of $N$ images as: 
\vspace{-0.5 em}
\begin{equation}
     \mathcal{L}_{\regPose} = \frac{1}{N}\sum_{i=1}^{N}\frac{1}{2{\regScale}_{i}^2}||{\gtPose}_i-\regPose_i||^2 +\frac{1}{2}log({\regScale}_{i}^2)
     \text{,}
     \label{eq:loss_kendall_pose}
\vspace{-0.5 em}
\end{equation}
where 
${\regScale}_i$ is the predicted per-sample variance from the \textit{scale network} ($\cscNN$), 
$\regPose_{i}$  is the predicted pose, and  
${\gtPose}_i$   is the ground-truth pose.  
The predicted pose and variance \textit{translates} and \textit{scales} the \bDF.

To minimize the loss in \cref{eq:loss_kendall_pose}, the network should infer a large variance, $\regScale$, when the predicted pose is far from the ground-truth one, and small otherwise. 
To discourage the network from always inferring a large $\regScale$ to naively minimise the loss, the second term of \cref{eq:loss_kendall_pose} acts as a regularizer.

Although the above formulation lets networks predict an \uncertainty estimate, \ie, the scale $\regScale_{i}$, we show in experiments that this does not correlate well with the pose error. 
Therefore, we need a more complex and ``expressive'' \bDF.

\zheading{\mbox{Sec. 3.2.2} Normalizing Flow (\NF)}
A Normalizing Flow (\NF) model \cite{realnvp} represents arbitrarily complex distributions. 
Thus, we extend the baseline model from above, and replace the Gaussian \bDF with a \NF network, $f_{\CNF}$; see ~\cref{fig:architecture}~\textcolor{red}{(b)}.
The \NF network, $f_{\CNF}$, transforms a simple distribution, $z \backsim \mathcal{N}(0,I)$, to a ``deformed'' zero-mean distribution $\bDFPose \backsim P_{\CNF}(\bDFPose)$, which is our new \bDF.
However, as with the Gaussian case, the \bDF is the same for all samples.
Since the \NF is an invertible model, let $\bDFPose = f_{\CNF}^{-1}(z)$. 
We then use a regressor, $\regressor$,
to infer pose, $\regPose_{i}$, and a per-sample variance, $\regScale_{i}$,
that \textit{shift} and \textit{scale} the \NF distribution $\bDFPose$ as:
\vspace{-0.35 em}
\begin{equation}
    \nfPose = \regPose + \bDFPose \regScale,
    \label{eq:transformBDF}
\vspace{-0.35 em}
\end{equation}
where we drop the image index $i$ for notational simplicity. 
Using \cref{eqn:1}, the \NF loss is defined as:
\vspace{-0.35 em}
\begin{equation}
    \small
    \begin{split}
    \mathcal{L_{NF}} 
    &
    = \left.-\textnormal{log}P_{\regressor,\CNF}(\nfPose|I)\right|_{\nfPose=\gtPose} 
    \\ & 
    = - \textnormal{log}P_{\CNF}(\gtDevPose) - \textnormal{log}\left|\textnormal{det }\frac{\partial f_{\CNF}^{-1} \left(\gtDevPose\right)}{\partial \gtPose}\right| 
    \\ & 
    = - \textnormal{log}P_{\CNF}(\gtDevPose) - \textnormal{log}\left|\textnormal{det }\frac{\partial (\gtPose-\regPose)/\regScale)}{\partial \gtPose}\right| 
    \\ &
    = - \textnormal{log}P_{\CNF}(\gtDevPose) + \textnormal{log}\regScale
    \text{,}
    \end{split}
    \label{eq:dle}
\vspace{-0.35 em}
\end{equation}
where $\gtDevPose = (\gtPose-\regPose)/\regScale$ is the scale-normalized error and the subscript $g$ denotes ground truth; 
note that for formulating the loss we set $\nfPose=\gtPose$, which is our goal.

\Cref{eq:dle} shows that, by using the \NF network, $f_{\CNF}$, training entirely depends on the estimation of $f_{\CNF}$; see the term $\textnormal{log}P_\CNF(\gtDevPose)$. 
We observe (as Li \etal~\cite{rle}) that this makes convergence challenging in early training stages, when both the regressor and the \NF module are untrained.

To tackle this problem, \rle~\cite{rle} proposes a gradient-shortcut approach. 
The solution is to introduce another simpler distribution, \ie, a Gaussian $Q(\gtDevPose)$, which is independent of the \NF module. 
As a result, the regressor network, $\regressor$, directly gets the gradients from the loss defined on $Q(\gtDevPose)$. 
The underlying assumption is that $Q(\gtDevPose)$ roughly matches the inferred-sample error distribution, while the \NF models a ``residual'' that is ``added'' on top of $Q(\gtDevPose)$ to eventually model a more complex distribution. 
Consequently, we modify the loss of \cref{eq:dle}, which becomes:
\begin{equation}
    \mathcal{L_{NF}} = - \textnormal{log}P_{\CNF}(\gtDevPose) - \textnormal{log}Q(\gtDevPose) + \textnormal{log}\regScale
    \text{.}
\end{equation}

\subsection{\DualCondFull (\DualCond)} \label{sec:cond-uncertain}
\modelname uses a novel \DualCondFull  to infer \uncertainty in a single feed-forward pass by augmenting the \bDF and \textit{scale network} ($\cscNN$) with a conditioning vector.

\zheading{\mbox{Sec. 3.3.1} Image-conditioned \bDF (\condnf)}
While a Normalizing Flow (\NF) can model arbitrarily complex distributions, having a single \bDF shared among all samples is a strong assumption. 
This assumption breaks when training on multiple data sources, as these can have different distributions; a shared \bDF struggles to model these. 
Even though the per-sample inferred pose, $\regPose$ and variance, $\regScale$, transform the base density function (\cref{eq:transformBDF}), we experimentally show that for complex scenarios like \HPS we need to go beyond a single shared \bDF. 
We argue that, by using additional per-sample information, a \bDF can represent complex scenarios.

Hence, we use a  conditional normalizing flow \cite{cond_nflow} 
(\condnf), $f_\CNF: \mathbb{R}^d \times \mathbb{R}^c \rightarrow \mathbb{R}^d$, to model the inferred-sample error for \HPS, where $d$ is the dimension of the \NF input and $c$ is the dimension of the conditioning vector; see \cref{fig:architecture}~\textcolor{red}{(c)}.
The flow model $f_\CNF$ is bijective in $z$ and $\bDFPose$, i.e. 
$z = f_{\CNF}(\bDFPose;\InputcnfNN)$ and $\bDFPose = f_{\CNF}^{-1}(z;\InputcnfNN)$, where $\InputcnfNN$ is the conditioning vector.
The \condnf takes some input vector $\InputcnfNN$ to model a different \bDF, the complexity of which depends on the discriminative power of the conditioning vector. 
A naive approach would be to use a one-hot encoding of the data source, to denote a different data distribution. 
However, for a rich dataset, even samples within it could belong to different distributions. 
Also, using a one-hot encoding as a conditioning vector limits training scalability. 

To account for this, we introduce image cues as the conditioning vector. 
The \emph{image features} are extracted by the regressor, $\regressor$, and  pass through a neural network, $\cnfNN$, which transforms them into a condition vector, $\InputcnfNN$. 
Our proposed conditional \NF loss is then: 
\begin{equation}
\small
        \mathcal{L_{NF}} = - \lambda_{\mathit{nf}}\textnormal{log}P_{\CNF}(\gtDevPose;\InputcnfNN) 
        - \lambda_{q}\textnormal{log}Q(\gtDevPose) 
        + \lambda_{\sigma}\textnormal{log}\regScale ,
\end{equation}
where $\lambda_{\mathit{nf}}$, $\lambda_{q}$ and $\lambda_{\sigma}$ are steering weights for each term.

\zheading{\mbox{Sec. 3.3.2} Pose-conditioned Scale (\condsc)}
The
\textit{scale network}, 
$\cscNN$, (\cref{fig:architecture}~\textcolor{red}{(c)}) infers \uncertainty as a scale, $\regScale$, that transforms a \bDF.
All related work conditions $\cscNN$ on \emph{image features} only. 
Here we argue that it is also important to determine \uncertainty based on \emph{pose plausibility} \wrt image evidence. 
We observe that pose provides a signal that helps $\cscNN$ regress \uncertainty that is highly correlated with reconstruction quality; this is critical: the pose and the image evidence should be consistent.

In the \modelname framework, we infer scale, $\regScale$, and \smpl pose, $\regPose$, through different sub-networks.
Then, $\cscNN$ concatenates pose with down-sampled image features, to create a balanced condition vector with same-dimension components (see \cref{sec:implementation_details} for details).
In the ablation experiments, we evaluate the importance of this novel formulation.

\zheading{\mbox{Sec. 3.2.3} Overall loss}
Given a regressor, we take its loss function and add $\mathcal{L_{NF}}$, as defined above.
With the current best practices~\cite{eft,pare} the overall loss is:
\vspace{-0.5 em}
\begin{equation}
    \begin{split}
        \mathcal{L} = \mathcal{L_{NF}} + \lambda_{\regShape}\mathcal{L}_{\regShape}(\regShape, \gtShape) + &\lambda_{3D}\mathcal{L}_{3D}(\threeDJnts, \gtThreeDJnts) + \\
        &\lambda_{2D}\mathcal{L}_{2D}(\twoDJnts, \gtTwoDJnts),
    \end{split}
\vspace{-0.5 em}
\end{equation}
where, 
$\mathcal{L}_{\regShape}$   is a \smpl shape loss, 
$\mathcal{L}_{\threeDJnts}$ is the \threeD joint loss and  $\mathcal{L}_{\twoDJnts}$   is the joint re-projection loss. 
The \twoD joints are calculated using the weak-perspective camera $\cam$ inferred by the regressor $\regressor$. 
The ground-truth \smpl shape is denoted by $\gtShape$, \threeD joints by $\gtThreeDJnts$ and \twoD joints by $\gtTwoDJnts$. 
Finally, $\lambda_{\regShape}$, $\lambda_{3D}$ and $\lambda_{2D}$ are steering weights for each term.

\vspace{-0.5em}
\section{Implementation details} \label{sec:implementation_details}

The \modelname framework is applicable to common \HPS regressors that infer parametric bodies like \smpl.
We show this by using three \HPS variants, \ie,
\hmreft \cite{eft}   (ResNet-50 \cite{resnet} backbone),
\pare   \cite{pare}  (HRNet-w32 \cite{hrnet}  backbone) and
\cliff  \cite{cliff} (HRNet-w48-cls \cite{hrnet} backbone).
We call the resulting methods \modelname-\-HMR-\-EFT, \modelname-\-\pare and \modelname-CLIFF, respectively. 

Image features, $\imageFeats$, of dimension 
$2048$, $3072$ and $2048$
are extracted after 
the global-feature pooling layer in \hmreft,
the part-attention aggregation layer in \pare, 
and the classifier layer in \cliff, 
respectively. 
The scale network $\cscNN$ is a \mbox{2-layer} MLP with 
$\{2048,216\}$ hidden layers. 
The first layer downsamples image features to a 216D vector before concatenating with the 216D ($24\times3\times3$) pose, to get a balanced condition vector.
The second layer infers a per-part \uncertainty (24D vector); for details and results on per-part \uncertainty, see \supmat

To get a single \uncertainty estimate for the full body, 
we sum the per-part uncertainties along \smpl's skeleton,  
normalize to the range of $[0,1]$, and compute the mean. 
A \mbox{1-layer} perceptron maps $\imageFeats$ into a 512D conditioning vector $\InputcnfNN$. 
\condnf is a \mbox{2-block} conditional RealNVP \cite{realnvp}; 
each block has 2 MLP layers of size $\{64,64\}$.
For details on the overhead of \modelname, see \supmat
We empirically set 
$\lambda_{\mathit{nf}} = 1.0e^{-4}$, 
$\lambda_{\sigma}      = 1.0e^{-4}$, 
$\lambda_{q}           = 1.0e^{-2}$; 
other $\lambda$ weights are as in original 
methods \cite{pare,eft,cliff}. 
We train with Adam~\cite{adam} ($3.0e^{-6}$ learning rate, 64 batch size).

\zheading{Training}
We train \modelname-\pare and \modelname-\hmreft on \coco \cite{coco}, \humanThreeSixFull \cite{human36m}, \mpiinf \cite{mpi-inf-3dhp}, \mpii \cite{mpii} and \lspet \cite{lspet}.
For datasets without 
\threeD ground truth (GT), we use \eft's \cite{eft} pseudo GT \smpl parameters, and use \hmreft's ratio for mixing \twoD and \threeD data in a batch.
Since \cliff's training code is not public, we re-implement it (``\emph{\cliffOur}''); for details see \supmat
For a fair comparison, we train with \cliff's~\cite{cliff} data and mixing ratio. 
For fast convergence, we start with a pre-trained \HPS 
~regressor,
and train the full \modelname model for  
50k iterations.
Then, we freeze the backbone and \HPS branch and train only the $\cscNN$ and \condnf nets for another 10k iterations.

For fair comparisons, for every benchmark setting we use the same training data and schemes as the original \HPS models, 
and use \pare's data augmentation. 

\zheading{Evaluation \& metrics}     \label{sec:evaluation_metrics}
We evaluate on the test sets of \threeDPW \cite{vonMarcard2018}, \threeDOH \cite{zhang3DOH} and \threeDPWOCC \cite{vonMarcard2018,zhang3DOH}. 
We report three \textit{pose metrics} ($mm$): 
the ``Procrustes aligned mean per joint position error'' (PA-MPJPE), 
the ``mean per joint position error'' (MPJPE) and 
the ``per-vertex error'' (PVE). 
We use the ``Pearson correlation coefficient'' (PCC) as the \textit{\uncertainty metric} to measure the \emph{correlation} of the estimated \emph{\uncertainty} with
\emph{pose error}.

\section{Evaluation}

\subsection{Performance of \modelname Framework} \label{sec:compare_sota}

\begin{table}
	\centering
	\resizebox{\columnwidth}{!}{
	\scriptsize
		\begin{tabular}{l|l|c|c|c|c}
			\toprule
			\bf Method & PVE $\downarrow$ & MPJPE $\downarrow$ & PA-MPJPE $\downarrow$ & Type \\
			\midrule
			HMMR \cite{hmr}                    & -& 116.5 & 72.6  & \deterM    \\
			VIBE \cite{vibe}                   & 113.4   & 93.5 & 56.5 & \deterM    \\
			Pose2Mesh \cite{pose2mesh}         & - & 89.2 & 58.9 & \deterM \\
			Zanfir \etal~\cite{pose2mesh}      & - & 90.0 & 57.1 & \deterM \\
			I2L-MeshNet \cite{l2l-meshnet}     & - & 93.2 & 58.6  & \deterM \\
			HMR \cite{hmr}                     & - & 130.0 & 76.7  & \deterM         \\
			SPIN \cite{spin}                   & 135.1 & 96.9  & 59.2  & \deterM    \\
			DSR \cite{dsr}                     & 105.8  & 91.7  & 54.1  & \deterM   \\
                HybriK~\cite{hybrik}$^\dagger$       & 86.5 & 74.1  & 45.0   & \deterM   \\

			Biggs \etal~\cite{multibodies} & - & 93.8 & 59.9 & \probM \\
			ProHMR \cite{prohmr}            & -    &  -      & 55.1 & \probM \\
			Sengupta \etal~\cite{hierprobhuman} & - & 84.9 & 53.6 & \probM \\
                HuManiFlow~\cite{humaniflow} & - & 83.9 & 53.4 & \probM \\
			\midrule
                \hmreft \cite{eft}                 & 106.1  & 92.5  & 54.2  & \deterM    \\
                \modelname-\hmreft                    &  101.1 &     88.5 &     52.4 & \probM   \\
			\textbf{\modelname-\hmreft-pGT}       & \bf 99.7 & \bf 87.3 & \bf 51.5 & \probM \\
                \midrule
                \PARE \cite{pare}                  & 97.9  & 82.0  & 50.9  & \deterM    \\
                \modelname-\pare                      &  95.3 &  80.3 &  49.9 & \probM \\
			\textbf{\modelname-\pare-pGT}         & \bf 94.0 & \bf 79.5 & \bf 49.4 & \probM \\
                \midrule
                \cliffOur~\cite{cliff}$^\dagger$    & 85.8 & 72.8  & 44.5  & \deterM    \\
                \modelname-\cliffOur$^\dagger$   &  84.6 &  70.9 &  43.3 & \probM \\
                \textbf{\modelname-\cliffOur-pGT}$^\dagger$   & \bf 83.5 & \bf 69.7 & \bf 42.8 & \probM \\
			\bottomrule
		\end{tabular} 
	}
	\vspace{-0.5 em}
	\caption{
	            \cheading{Evaluation of \modelname \& SOTA \HPS
	            on \threeDPW (\cref{sec:compare_sota})}
	            \captionUnits. 
                    \cliffOur is our re-implementation of CLIFF~\cite{cliff}, and $^\dagger$ denotes that \threeDPW is used for training. 
                    Suffix ``-pGT'' denotes self-improved training for several \modelname variants (see \cref{sec:downstream_application}, \mbox{task 1}).
                    ``\deterM'' and ``\probM'' refers to deterministic and probabilistic methods, respectively.
        }
        \label{tab:compare_sota_3dpw}
	\vspace{-1.0 em}
\end{table}
\begin{table}
	\centering
	\scriptsize
	\resizebox{\columnwidth}{!}{
		\begin{tabular}{l|c|c|c|c|c}
			\toprule
			& \multicolumn{3}{c|}{ \threeDPWOCC \cite{zhang3DOH,vonMarcard2018} } & \multicolumn{2}{c}{ \threeDOH \cite{zhang3DOH} } \\
			\cmidrule(lr){2-6}
			\textbf{Method} & {PVE $\downarrow$} & {MPJPE $\downarrow$} & {PA-MPJPE $\downarrow$} & {MPJPE $\downarrow$} & {PA-MPJPE $\downarrow$} \\
			\midrule
			Zhang~\etal~\cite{zhang3DOH}     &     -      &     -    &     72.2 &      -    &      58.5 \\
			SPIN \cite{spin}        &     121.6  &     95.6 &     60.8 &     104.3 &      68.3 \\
			\PARE \cite{pare}        &     111.3  &     90.5 &     56.6 &      63.3 &      44.3 \\
			\midrule
			\textbf{\modelname-\pare}     & \bf 109.1  & \bf 89.0 & \bf 54.5 &  \bf 61.0 &  \bf 42.5 \\
			\bottomrule
		\end{tabular} 
	}
	\vspace{-0.5 em}
	\caption{
	            \cheading{Evaluation on occlusion datasets (\cref{sec:compare_sota})}
	            \captionUnits. 
	            All methods use a ResNet-50 baseline. 
	}
	\label{tab:compare_sota_occ}
	\vspace{-1.2 em}
\end{table}

We compare recent \HPS models and \modelnameHPS (building on three recent \HPS models: \hmreft, \pare, \cliff) on \threeDPW's test set in \cref{tab:compare_sota_3dpw}
We list two method types: 
(1) ``deterministic'' ones (\deterM), inferring \smpl parameters, and 
(2) ``probabilistic'' ones, inferring a distribution either over \smpl parameters or sample error in case of \modelnameHPS.
The most direct comparison is with \hmreft, \pare and \cliff that use the same architecture and training; 
\modelnameHPS outperforms its base \HPS model.
Note that 
some probabilistic methods (\ie, ProHMR) infer 
multiple bodies, but \modelname infers only one.

We also evaluate on occlusion datasets (\threeDPWOCC \cite{zhang3DOH,vonMarcard2018} and \threeDOH \cite{zhang3DOH}) in \cref{tab:compare_sota_occ}. 
We train \modelname-\pare with a ResNet-50 baseline as all other models. 
All models but SPIN train on \coco, \humanThreeSixFull and \threeDOH. 
\modelname-\pare performs best; 
our framework helps improve \HPS.

We qualitatively compare \modelnameHPS with its ``deterministic'' \HPS model (\pare \cite{pare}, CLIFF \cite{cliff})
and ``probabilistic'' \HPS models (Sengupta \etal~\cite{hierprobhuman}, ProHMR \cite{prohmr}) in \cref{fig:uncert_compare}.
The former do not represent \uncertainty (bodies shown in gray color), while for the latter, \uncertainty is visualized color-coded (see bar in \cref{fig:uncert_compare}). 
Sengupta \etal~sample 64 \smpl bodies and compute \uncertainty as the per-vertex variance. 
For ProHMR we do the same sampling, but compute \smpl-pose variance and take the mean along \smpl's skeleton as \uncertainty, like \modelname does (see \cref{sec:implementation_details}).
\modelnameHPS outperforms its base \HPS model in challenging cases.
ProHMR infers similar \uncertainty values across images (see consistent ``aquamarine'' color), irrespective of image ambiguities or the inferred pose quality. 
Sengupta \etal~consistently infer high uncertainty for end-effectors in a narrow range, having an inductive bias. 
\modelnameHPS infers better \smpl bodies, and an \uncertainty that is better correlated with its output quality.

\subsection{Performance of \Uncertainty Formulations} \label{sec:compare_poco_baselines}

We compare the performance of our \uncertainty formulation (\ie, our \textit{\DualCondFull}) with existing \uncertainty formulations~\cite{kendall_uncert, rle} on \threeDPW~\cite{vonMarcard2018}; see \cref{tab:ablation}.
We use \hmreft~\cite{eft} as the \HPS model to evaluate the performance of each formulation; for other \HPS models see \supmat
All models are trained on the same data.

We adapt the \uncertainty formulation of Kendall \etal \cite{kendall_uncert} (\textit{\BLGauss}) and RLE~\cite{rle} (\textit{\BLNFlow}) for \HPS (2nd and 3rd row, respectively, in \cref{tab:ablation}).
Following their approach, we use a Gaussian distribution in \textit{Gauss} and use normalizing flow (\NF) in \textit{\BLNFlow}.
The pose metrics slightly improve when the network is forced to estimate \uncertainty.
However, with \modelname's \uncertainty formulation (5th row in~\cref{tab:ablation}), the pose metrics and the \uncertainty metric (PCC) improve significantly compared to prior methods; depicting the efficacy of our framework.
The PCC metric shows how uncertainty correlates with pose error; a higher value denotes a stronger correlation. %
\vspace{+0.7 em}

\subsection{Performance of \modelname Components}
\label{sec:poco_ablations}
\modelname introduces a novel \textit{\DualCondFull} which consists of \textit{\condnf} (\cref{sec:cond-uncertain}\textcolor{red}{.1}) and \textit{\condsc} (\cref{sec:cond-uncertain}\textcolor{red}{.2}).
We evaluate the performance of each component; see 4th and 5th row of~\cref{tab:ablation}.
For \textit{\condsc}, we use \NF as \bDF and condition scale inference (via $\cscNN$) on pose, and get further improvement.
For \textit{\condnf}, we realize that images can look very different, even within the same dataset, and cannot all share the same \bDF. Intuitively, images that lie close-by in a feature space can share a \bDF, distant ones can not.
We add image features as (via $\cnfNN$) for the \NF network; this also boosts performance.

The \textit{\DualCondFull} of our \modelname framework (6th row of~\cref{tab:ablation}) results in the best performance for all metrics, showing that the above two design choices contribute positively and are complementary.
Comparing \modelname with pure-\HPS models (first row of \cref{tab:ablation}) shows a notable difference.
Perhaps surprisingly, as this was not our goal, pushing our model to solve a harder task, \ie, perform both \HPS and \uncertainty inference, improves \HPS performance; this 
aligns with findings in multi-task learning~\cite{taxonomy}.

\begin{table}[t]
	\centering
	\resizebox{\columnwidth}{!}{
	\scriptsize
		\begin{tabular}{l|c c c c}
			\toprule
			\bf POCO Variants & PVE $\downarrow$ & MPJPE $\downarrow$ &  PA-MPJPE $\downarrow$ & PCC $\uparrow$ \\
			\midrule
			\hmreft~\cite{eft}              &     106.1 &     92.5 &     54.2 &       -      \\
			\hphantom{l}+ \BLGauss~\cite{kendall_uncert}          &     105.7 &     92.3 &     54.1 &       0.31   \\
			\hphantom{l}+ \BLNFlow~\cite{rle}          &     104.9 &     91.2 &     53.7 &       0.42   \\
			\hphantom{l}+ \condsc         &     103.5 &     89.9 &     53.3 &       0.44   \\
			\hphantom{l}+ \condnf         &     103.2 &     89.8 &     53.1 &       0.46   \\
			\hphantom{l}+ \modelname        &     101.1 &     88.5 &     52.4 &       0.52   \\
			\cmidrule(lr){1-5}
			\modelname-\hmreft-pGT    & \bf  99.7 & \bf 87.3 & \bf 51.5 &  \bf  0.53   \\
			
			\bottomrule
		\end{tabular} 
	}
        \vspace{-0.8 em}
	\caption{
                    \cheading{Evaluation of \modelname (\cref{sec:poco_ablations}) and other \uncertainty methods on \threeDPW (\cref{sec:compare_poco_baselines})} 
	            PCC is in the range $\in[-1, 1]$. 
                    We evaluate on \threeDPW's test set \cite{vonMarcard2018}; 
                    no model was trained on \threeDPW.
                    Suffix ``-pGT'' denotes self-improved training (\cref{sec:downstream_application}, \mbox{task 1}).
                    The row ``+\modelname'' denotes the model ``\modelname-\hmreft''. 
	}
    \label{tab:ablation}
    \vspace{-1.3 em}
\end{table}
\begin{figure*}
    \vspace{-1.0 em}
    \centerline{\includegraphics[width=\textwidth]{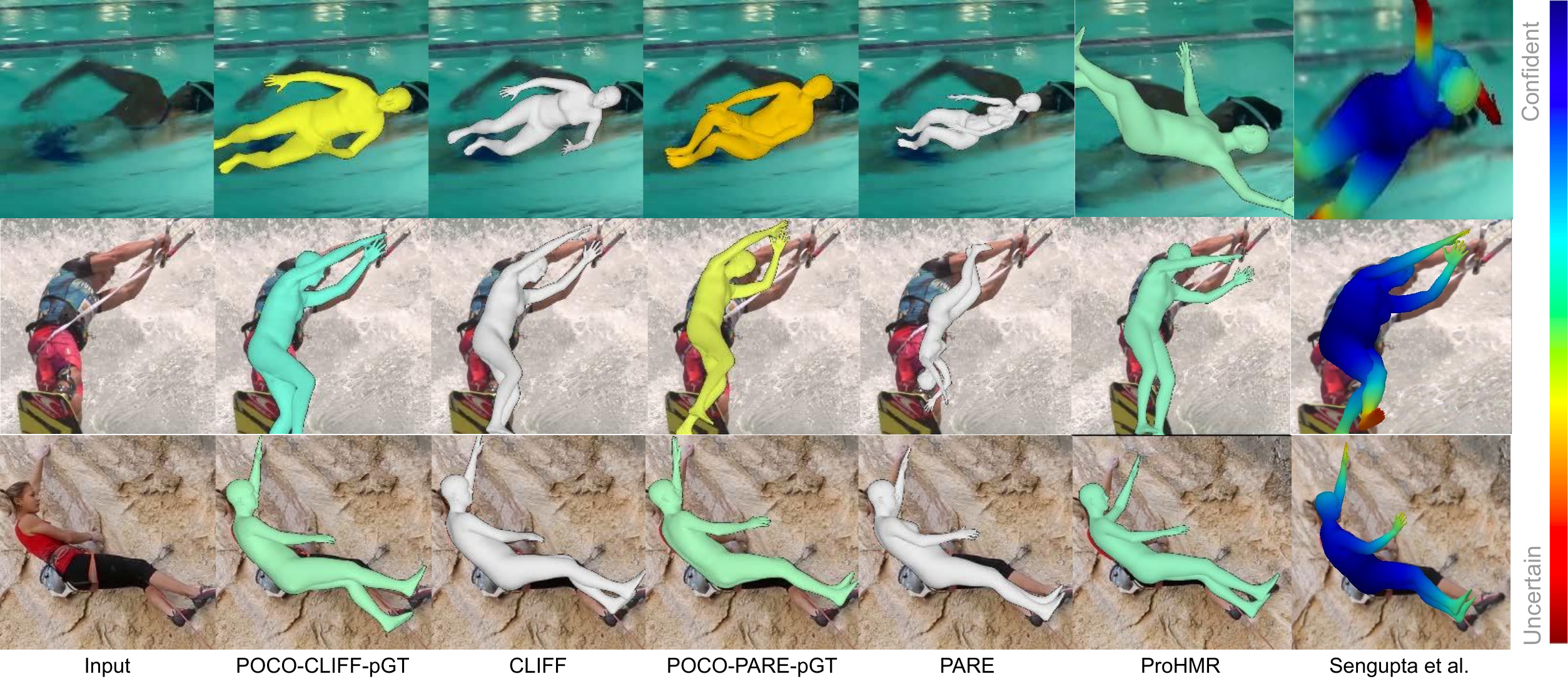}}
    \vspace{-0.5 em}
    \caption{
        \cheading{Qualitative results for \inthewild images}
        We evaluate  
        \cliffOur   \cite{cliff},
        \PARE       \cite{pare}, 
        ProHMR      \cite{prohmr},
        Sengupta    \etal~\cite{hierprobhuman} and 
        our \modelname built on two different \HPS regressors.
        For more results see \supmat
    }
    \vspace{-1.5 em}
    \label{fig:uncert_compare}
\end{figure*}

\begin{figure}
    \centerline{\includegraphics[width=\linewidth]{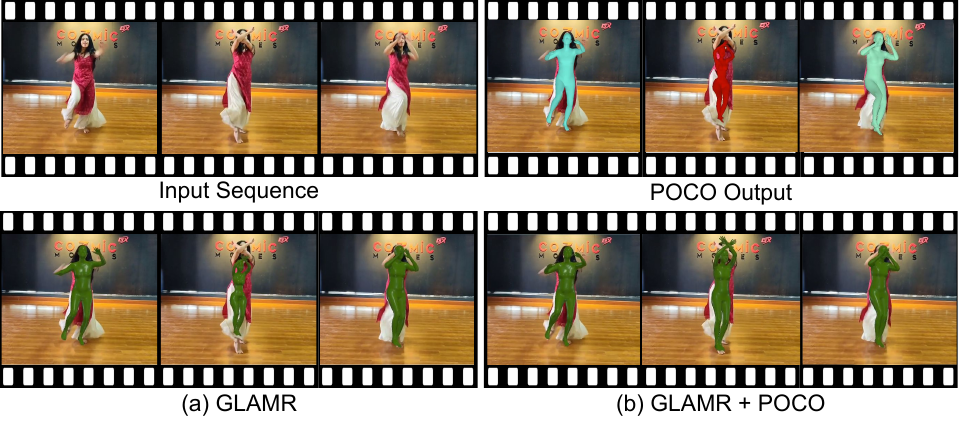}}
    \vspace{-0.3 em}
    \caption{
                \cheading{Infilling with \uncertainty}
                Given estimated 3D bodies for detected people in a video, GLAMR~\cite{yuan2021glamr} infills 3D bodies for missing detections, and does sequence-level refinement for these. 
                (a) 
                However, GLAMR considers all estimated 3D bodies as accurate, and fails when they are not. 
                (b) 
                \modelname provides both 3D bodies and an uncertainty metric; 
                this helps GLAMR reject uncertain bodies and infill 
                also for these.
                See middle frame for a/b.
                For video results, see the \video.
    }
    \label{fig:infilling}
    \vspace{-1.0 em}
\end{figure}

\subsection{Downstream Tasks}   \label{sec:downstream_application}
We show the usefulness of \modelname's \uncertainty for two downstream applications, as discussed in the following. 

\zheading{Task 1: Self-improved} \HPS Training
To train robust \HPS regressors, 
we need highly-varied images paired with high-quality \threeD bodies. 
Datasets with \threeD ground truth (GT) are small and captured in %
lab settings, thus, recovering good \pseudoGT \cite{eft} from \inthewild images has proven to be helpful.
However, this typically requires human annotators to manually curate good \HPS reconstructions \cite{eft}.

We automate this with \modelname. 
We use the model's \uncertainty measure to recover \threeD bodies from Charades \cite{charades}, a dataset with in-the-wild videos of daily activities. 
By using \modelname's \uncertainty, which is correlated with its output quality, we automatically curate the \smpl estimates with low \uncertainty; 
we use a strict \uncertainty threshold of $0.3$ and keep only estimates below it; for details see \supmat 
We treat the curated bodies as \pseudoGT, add this to the existing training data, and fine-tune \modelname variants; these self-improved models are denoted with the \mbox{``-pGT''} suffix.

We evaluate the self-improved models on \threeDPW; see bottom of \cref{tab:compare_sota_3dpw,tab:ablation}, \mbox{``-pGT''} entries.
Unsurprisingly, they improve all error metrics (PVE, MPJPE). 
In \cref{tab:ablation} the self-improved model has the same correlation (PCC) between \uncertainty and output quality. 
We hypothesize that \pseudoGT contributes highly-varied images that improve \HPS, but the strict \uncertainty threshold contributes less-varied \uncertainty labels. 
We also vary the \uncertainty threshold and observe that for higher thresholds, the performance decreases, showing that higher \uncertainty corresponds to low quality pseudo-GT; for details see \supmat

\zheading{Task 2} 
\fourD bodies from videos
To recover \fourD bodies from videos, methods like \glamr \cite{yuan2021glamr} infer per-frame \smpl bodies via an \HPS model, and post-process them to improve temporal quality, while ``infilling''  frames with missing detections. 
However, existing methods \emph{cannot automatically curate} the ``good" \smpl estimates, consequently, results may be ``contaminated'' with \HPS failures.

We address this with \modelname. 
First, we run \modelname on a sequence and estimate per-frame bodies along with \uncertainty. 
We use a threshold of $0.8$, discard estimates with \uncertainty values above this, and consider these rejected frames as ``missing''. 
We then use \glamr's infiller to generate the missing bodies and perform global optimization for refinement. 
By rejecting the \uncertain bodies, the infiller has to work harder but relies on neighboring high-quality poses.
As seen in \cref{fig:infilling}, \modelname helps recover accurate \HPS sequences for videos using its \uncertainty measure. 
In such scenarios, standard \HPS models have a clear disadvantage.
The above shows that \modelname's \HPS performance and \uncertainty inference are promising and useful.

\section{Conclusion}
While a huge progress has been made in estimating \threeD human bodies from an image, most regressors do not have a measure for the output's quality. 
Thus, downstream applications do not know how much to rely on them, and struggle with bad outputs. 
We account for this with \modelname, a novel framework that infers \smpl bodies along with its \uncertainty.
Experiments show that \modelname's \uncertainty is correlated with the quality of pose reconstruction.
We also show that \modelname can build on three different \HPS models; {this is promising} for also using future ones. 
Finally, we show the usefulness of \uncertainty for 
practical applications.

\qheading{Future work}
\modelname models pose \uncertainty; 
this can be extended also for shape; for discussion see \supmat 
For reconstructing challenging videos in \fourD, methods like GLAMR~\cite{yuan2021glamr} can benefit from \modelname's \uncertainty estimate. We think that \modelname will probe further work on \uncertainty for \HPS.

\pagebreak

{\bf Acknowledgements:}
We sincerely thank Partha Ghosh and Haiwen Feng for insightful discussions and Priyanka Patel for CLIFF reference implementation. 
We also thank Peter Kulits, Shashank Tripathi, Muhammed Kocabas and rest of Perceiving Systems department for their feedback. 
The authors thank the International Max Planck Research School for Intelligent Systems (IMPRS-IS) for supporting Sai Kumar Dwivedi.
This work was partially supported by the German Federal Ministry of Education and Research (BMBF): Tübingen AI Center, FKZ: 01IS18039B.

MJB has received research gift funds from Adobe, Intel, Nvidia, Meta/Facebook, and Amazon. MJB has financial interests in Amazon, Datagen Technologies, and Meshcapade GmbH. While MJB is a part-time employee of Meshcapade, his research was performed solely at, and funded solely by, the Max Planck Society.

{
    \small
    \bibliographystyle{config/ieeenat_fullname}
    \bibliography{paper}
}

\clearpage


\appendix

\setcounter{figure}{0} \renewcommand{\thefigure}{S.\arabic{figure}}
\setcounter{table}{0} \renewcommand{\thetable}{S.\arabic{table}}

\begin{strip}%
 \centering \Large
 \textbf{%
  \modelname: \threeD Pose and Shape Estimation with Confidence \\
 \vspace{0.5cm} \emph{**Supplementary Material**} \vspace{0.5cm}
 }\\
\end{strip}

In this Supplementary-Material document, we provide more details about our method.
Additionally, please see the \video for a summary of the method and more visualizations of the results.

\section{Regressor Network Architecture}

We use three variants of \HPS regressors in \modelname, \ie, \pare~\cite{pare}, \hmreft~\cite{eft},
\cliff~\cite{cliff} as shown in ~\cref{fig:supmat_regressor_arch}.

In \pare, the input image is first passed through a CNN backbone (HRNet-32W), 
and features are extracted before the average pooling layer. 
The features are then passed through two separate branches: a \emph{\twoD Part Segmentation} branch and a \emph{\threeD Body Feature} branch. 
The \twoD part segmentation branch produces body-part attention features $S \in \mathbb{R}^{H \times W \times (J+1)}$, where $J=24$ is the number of \smpl body parts, while a background mask is assigned to non-human pixels. 
The body feature branch is used to estimate \smpl body parameters. 
Both branches produce features of the same spatial dimensions, $H \times W$. 
The features from $S$ pass through a spatial softmax normalization layer, $\kappa$. 
These are used as soft attention masks to aggregate \threeD body features into final features, %
$F = \kappa(S)^{\top} \odot B$, where $S \in \mathbb{R}^{HW \times J}$, $B \in \mathbb{R}^{HW \times C}$ and $F \in \mathbb{R}^{J \times C}$; 
note that $S$ and $B$ are reshaped before the operation. 
Each feature row, $F_{i} \in \mathbb{R}^{1 \times C}$ with $i \in \{ 1, \dots, J \}$, 
passes through a separate MLP to get \smpl pose parameters, $\theta = \{ \theta_i \}$. 
To estimate the camera, $C$, and \smpl shape, $\beta$, all final features $F$ are fed, concatenated, to different MLPs.

\begin{figure}[t!]
    \centerline{\includegraphics[width=1.00 \linewidth]{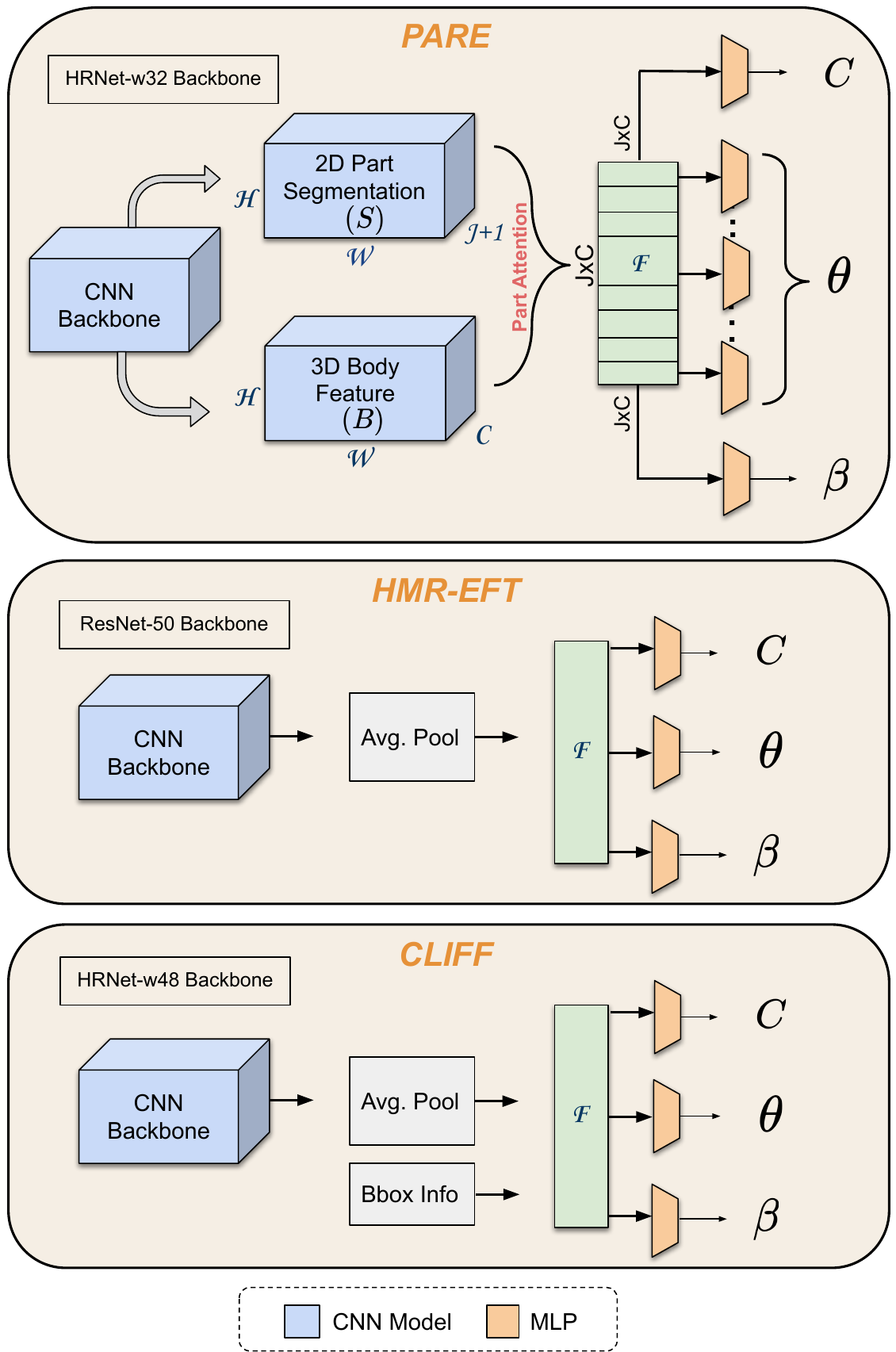}}
    \caption{
                \cheading{Regressor architecture}
    }
    \label{fig:supmat_regressor_arch}
\end{figure}

\hmreft uses a simple network architecture for estimating \HPS. The input image passes 
through a CNN backbone (ResNet-50) followed by a global average pooling layer. The features from the pooling layer are used to regress \smpl pose, $\theta$, shape, $\beta$, and camera parameters, $C$, through separate MLPs. 
This regression is done through an iterative error feedback loop.

\cliff uses a HRNet-w48 network architecture as a CNN backbone. Along with the a cropped image, \cliff takes the bounding box location information (Bbox Info) as input to provide the location information of the person in the image. 
This helps to accurately predict the global rotation in the original camera coordinate frame. The bounding box formation contains the center of bounding box center relative to image center and focal length of the original camera which is calculated using image height and width.
Contrary to \pare and \hmreft, \cliff computes a 2D keypoint loss after projecting the body keypoints onto the original image plane.

\section{\cliff Training and Evaluation}

Since the \cliff training code is not public, we re-implement it (``\emph{\cliffOur}"). We train \cliffOur on \coco~\cite{coco}, \mpii~\cite{mpii}, \mpiinf~\cite{mpi-inf-3dhp}, \humanthreesix~\cite{human36m} and \threeDPW~\cite{vonMarcard2018} with the same dataset ratios used in \hmreft~\cite{eft}. 
For \twoD datasets, we use the pseudo \groundtruth \smpl parameters provided by \cliff and, for other datasets, we use the the original annotations provided by the respective datasets. 
Following prior work~\cite{pare, eft}, we resize the cropped image to $224 \times 224$ for both training and evaluation.
To compute the 2D keypoint loss on full image plane, first we crop the keypoints according to the person bounding box and then project them back to the original image size.
To evaluate on \threeDPW test, we use the same bounding box scale and center used by prior work~\cite{pare,spin}.
\begin{table}[t]
    \centering
    \resizebox{\columnwidth}{!}{
        \scriptsize
        \begin{tabular}{l | c c | c c | c c |}
            \toprule
            \bf \multirow{2}{*}{Method} & \multicolumn{2}{c|}{ \hmreft~[\textcolor{green}{17}] } & \multicolumn{2}{c|}{ \pare~[\textcolor{green}{26}] } & \multicolumn{2}{c|}{ \cliff~[\textcolor{green}{37}] } \\
            \cmidrule(lr){2-7}
                & PVE $\downarrow$ & PCC $\uparrow$ &  PVE $\downarrow$ & PCC $\uparrow$ & PVE $\downarrow$ &  PCC $\uparrow$ \\
            \midrule
            \BLdeterministicHPS                              &     106.1  &     -   &      97.9  &     -     &    85.8 &   -    \\
            \BLGauss~[\textcolor{green}{22}]   &     105.7  &    0.31 &      97.1  &     0.32 &     85.4 &   0.29  \\
            \BLNFlow~[\textcolor{green}{35}]   &     104.9  &    0.42 &      96.6  &     0.44 &     85.3 &   0.40   \\
            \midrule
            \bf \modelnameHPS    &  \bf 101.1  &  \bf 0.52 &   \bf 95.3  &  \bf 0.54 &  \bf 84.6 &    \bf 0.51   \\
            
            \bottomrule
        \end{tabular}
    }
    \caption{\small{Evaluation of \modelname and other \uncertainty formulations for different \hps regressors.}
    }
    \label{tab:supmat_pcc}
\end{table}

\section{Performance of Uncertainty Formulations for different \HPS regressors}

We compare the performance of our uncertainty formulation (i.e., our \DualCondFull) with existing uncertainty formulations~\cite{kendall_uncert, rle} on \threeDPW~\cite{vonMarcard2018} for different \HPS regressors~\cite{hmr, pare, cliff} as shown in~\cref{tab:supmat_pcc}. This complements  Tab.~\textcolor{red}{3} in the main paper.
Our \uncertainty formulation outperforms the prior formulations (\textit{\BLGauss} and \textit{\BLNFlow}) for all \HPS regressors in both the pose (PVE) and \uncertainty (PCC) metrics.
Note that the PCC metric should not be compared across different \HPS methods on its own.
Focusing \textit{separately} on each \HPS method, the important thing is that our novel \uncertainty formulation %
consistently lowers PVE errors while increasing the PCC metric; this is indicative of a better \uncertainty formulation.

\section{Per-Part and Per-Vertex \Uncertainty}

\begin{figure}[t!]
    \centerline{\includegraphics[width=\linewidth]{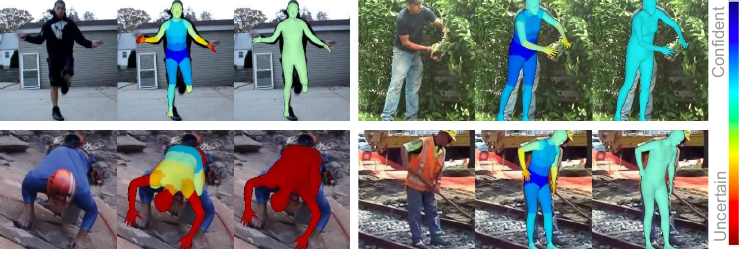}}
    \caption{\cheading{Per-part uncertainty of \modelname-\cliff}
    For each image triplet: input image, per-part uncertainty of \modelname and whole-body uncertainty of \modelname. 
    }
    \label{fig:supmat_per_part_uncertainty}
\end{figure}

\modelname models the \uncertainty of \smpl pose parameters in the following way. 
First, it estimates the \uncertainty for the axis-angle rotation of each of \smpl's skeleton joints separately. 
This is important because each of these has a different amount of error.
However, for downstream applications, having a single \uncertainty value for the full body is more practical. 
To this end, we traverse \smpl's kinematic chain (\ie, recursively going in the direction from parent to child), and add the axis-angle uncertainties of the respective skeleton joints; as there 24 joints in total, this produces a 24D vector. 
We then normalize the 24D vector to the range of $[0,1]$ and compute the mean to get a single scalar \uncertainty value; this represents the \uncertainty for the full body.
The per-part \uncertainties and the full-body \uncertainty are shown in ~\cref{fig:supmat_per_part_uncertainty}.

A few recent methods~\cite{hierprobhuman, humaniflow} show per-vertex \uncertainties. They do so by sampling multiple bodies and computing \uncertainty as per-vertex variance. While this is an interesting choice, modelling per-vertex \uncertainties in a single feed-forward pass would be expensive. One would need to model the \textit{base density function} and \textit{scale network} to output $6890$ (\smpl vertices) as compared to only $24$ variables (\smpl joints) in the case of \modelname.

\section{Overhead of \modelname Framework}

\begin{table}[t]
    \centering
        \scriptsize
        \begin{tabular}{l|c c c }
            \toprule
            \bf Method                &   Train-Params &    Test-Params &   Inference Time  \\                     
            \midrule
            \cliff                    &       81.0 M       &    81.0 M     &      1.45 ms  \\
            \modelname-\cliff         &       82.6 M    &       81.3 M    &       1.49 ms    \\		
            \bottomrule
        \end{tabular}
    \caption{\small{\modelname's overhead when applied to \hps regressor.}
    }
    \label{tab:sup_overhead}
    \vspace{-1 em}
\end{table}

\modelname is a general \uncertainty framework that can be applied to common \HPS methods, extending them to also estimate \uncertainty.
It adds a \textit{\bDF} and \textit{scale network} for estimating \uncertainty in a single network pass.
\cref{tab:sup_overhead} shows that \modelname imposes only a small overhead.
\modelname-\cliff has only 2\% more training parameters than \cliff due to adding the \bDF and scale network.
The former is unused at test time and the latter is just a small NN, so, adding \modelname increases inference time only minimally.

\begin{figure}[t!]
    \centerline{\includegraphics[width=\linewidth]{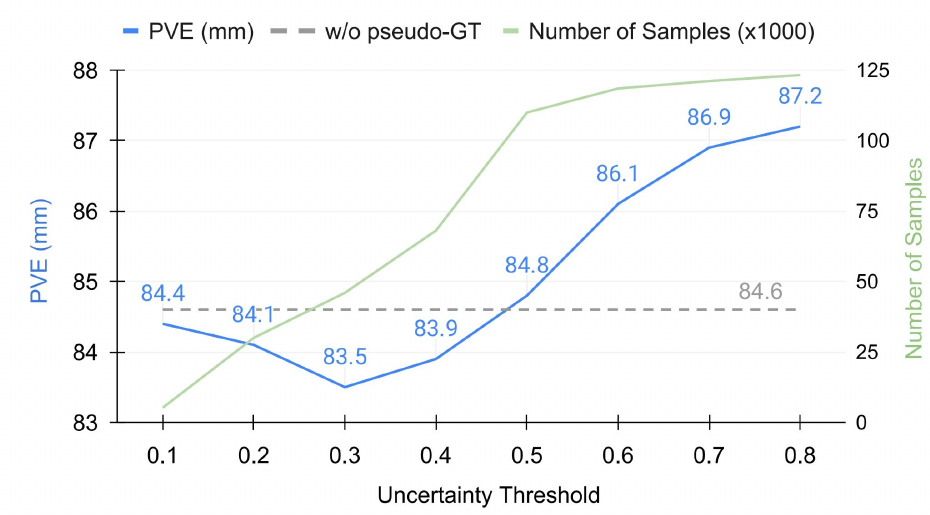}}
    \vspace{-0.1in}
    \caption{
        \cheading{\Uncertainty threshold for pseudo-GT selection}
        Using an optimal \uncertainty threshold of $0.3$ for selecting pseudo-GT from Charades dataset~\cite{charades} for training, \modelname-\cliff-pGT's performance on \threeDPW test set is better than the model trained without it (dashed line). The performance degrades with higher uncertainty threshold. PVE denotes per-vertex error. The green line is for number of samples and axis is on the right.
    }
    \label{fig:plot_augment_TrainingData_w_POCO_annotator}
\end{figure}

\section{Self-Improved \HPS Training}

\modelname estimates an \uncertainty measure that correlates with pose reconstruction quality.
We use this measure to automatically curate \smpl estimates from the Charades dataset~\cite{charades} and improve \modelname, 
using the following steps.

We first sample 
every 100th frame from the videos to get a total of $130$K images, and apply \modelname on these. 
We then vary \modelname's \uncertainty threshold, and for each value we 
automatically curate the produced \smpl estimates and extend \modelname's training data. 
This results in multiple extended training datasets.
We finetune \modelname separately for each of these, and evaluate each finetuned model on \threeDPW.

The evaluation results are shown in 
\cref{fig:plot_augment_TrainingData_w_POCO_annotator}. 
The dashed gray line 
shows the original \modelname (with no additional pseudo-GT). %
The blue  curve shows the PVE error (mm) of the finetuned variants.
The green curve shows the number of curated samples for each threshold.
With a low threshold ($0.1$) very few samples pass, 
thus, performance is almost unchanged. 
With a high threshold ($\geq 0.45$), as the threshold gets higher, more samples of decreasing quality pass, which can even harm performance. 
For thresholds in the range of $\bigl[ 0.2, 0.4 \bigr]$ enough good-quality samples pass so that performance improves. 
The best performance is achieved for a threshold of $0.3$,
which results in adding roughly 
46k 
samples in the training dataset; 
given the limited number of subjects and pose variation compared to the original dataset, the performance shows 
that this bootstrapping is promising.
Note that the threshold is determined on the Charades dataset by visual inspection. \threeDPW test data is not used in setting the threshold.
Random samples of pseudo \groundtruth generated by \modelname-\cliff on
Charades dataset is shown in ~\cref{fig:supmat_charades_pGT}.

\begin{table}[t]
    \centering
    \resizebox{\columnwidth}{!}{
    \scriptsize
    \begin{tabular}{l|c c c c c}
        \toprule
        \bf Method                   & PVE $\downarrow$ & MPJPE $\downarrow$ &  PA-MPJPE $\downarrow$ &  Filter pGT & \# pGT \\
        \midrule
            \modelname-\cliff                    &       84.6    &      70.9    &       43.3  & - & -  \\
            \modelname-\cliff-Whole        &       87.2    &       74.7    &       46.1  & \xmark & 130K \\
            \modelname-\cliff-Rand       &       86.6    &       73.9    &       45.6  & \xmark & 46K\\
        \bf \modelname-\cliff-pGT        &   \bf 83.5    &   \bf 69.7    &   \bf 42.8  & \cmark  & 46K\\		
        \bottomrule
    \end{tabular}
    }
    \vspace{-0.5 em}
    \caption{
        \textbf{Effect of uncertainty-filtered pGT data on \threeDPW.}
        ``Whole'' trains with all data~\cite{charades} without filtering, 
        ``Rand'' with random samples, and ``pGT'' filters using POCO \uncertainty.
    }
    \vspace{-1.0 em}
    \label{tab:sup_unsupervised}
\end{table}

To better understand the value of self-improvement, we perform two additional baseline experiments.
\modelname-\cliff-pGT uses 46K frames (out of 130K) from the Charades dataset, filtered using our \uncertainty estimates.
For comparison, we re-train \modelname-\cliff: 
(1) using all 130K frames   \emph{(``Whole'')}, and 
(2) using 46K frames randomly sampled from the 130K  \emph{(``Rand'')}.
All methods use  \modelname-\cliff \smpl estimates as pGT.
\cref{tab:sup_unsupervised} shows that  %
adding data without confidence filtering makes results worse, while our self-improvement process 
improves them.

\begin{figure}[t!]
    \centerline{\includegraphics[width=\linewidth]{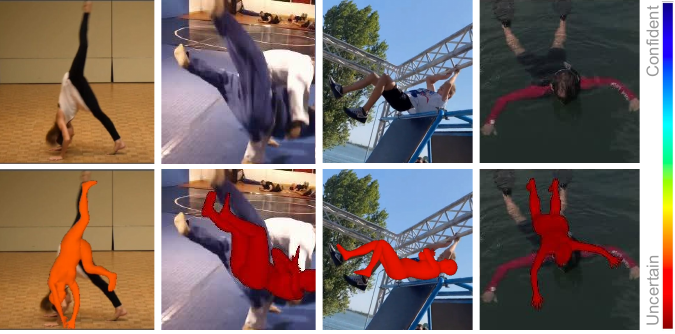}}
    \vspace{-0.1in}
    \caption{\cheading{Failure cases of \modelname}
    In some cases of occlusion and \ood poses, \modelname estimates high uncertainty even though the pose reconstructions are not totally implausible. 
    }
    \label{fig:supmat_failure_case}
    \vspace{-0.5em}
\end{figure}

\section{Failure Cases}

In Fig.~\cref{fig:supmat_failure_case}, we show some representative cases in which \modelname's prediction quality and its uncertainty estimate disagree.
Typically, \modelname produces more plausible poses than other \HPS methods~\cite{pare,hier_mesh_recovery}, even for complex scenarios of heavy occlusion and \ood poses. 
However, sometimes \modelname estimates high \uncertainty, even if the poses it produces are reasonable; think of this a ``false negative''.
In \cref{fig:supmat_failure_case} each image either contains an unusual pose, motion blur, occlusion, or dim lighting -- in some cases more than one of these.
It is reasonable for the network to be uncertain of its estimates in these cases, even if it happens to get the pose right (or close).

\section{Effect of Occlusion on \Uncertainty}

\modelname estimates \threeD body parameters and their uncertainty in a single feed-forward pass.
The \uncertainty is correlated to image ambiguities and the quality of reconstruction.
We analyze the correlation  qualitatively on \threeDPW for the \modelname-\hmreft network. 
Specifically, we add a synthetic occluder that we swipe throughout the video frames to see the effect on \uncertainty; see \cref{fig:supmat_uncertainty_analysis}. 
We observe that \uncertainty increases when a body part is occluded.
\section{Qualitative Results}

\begin{figure*}
    \centerline{\includegraphics[width=1.0\textwidth]{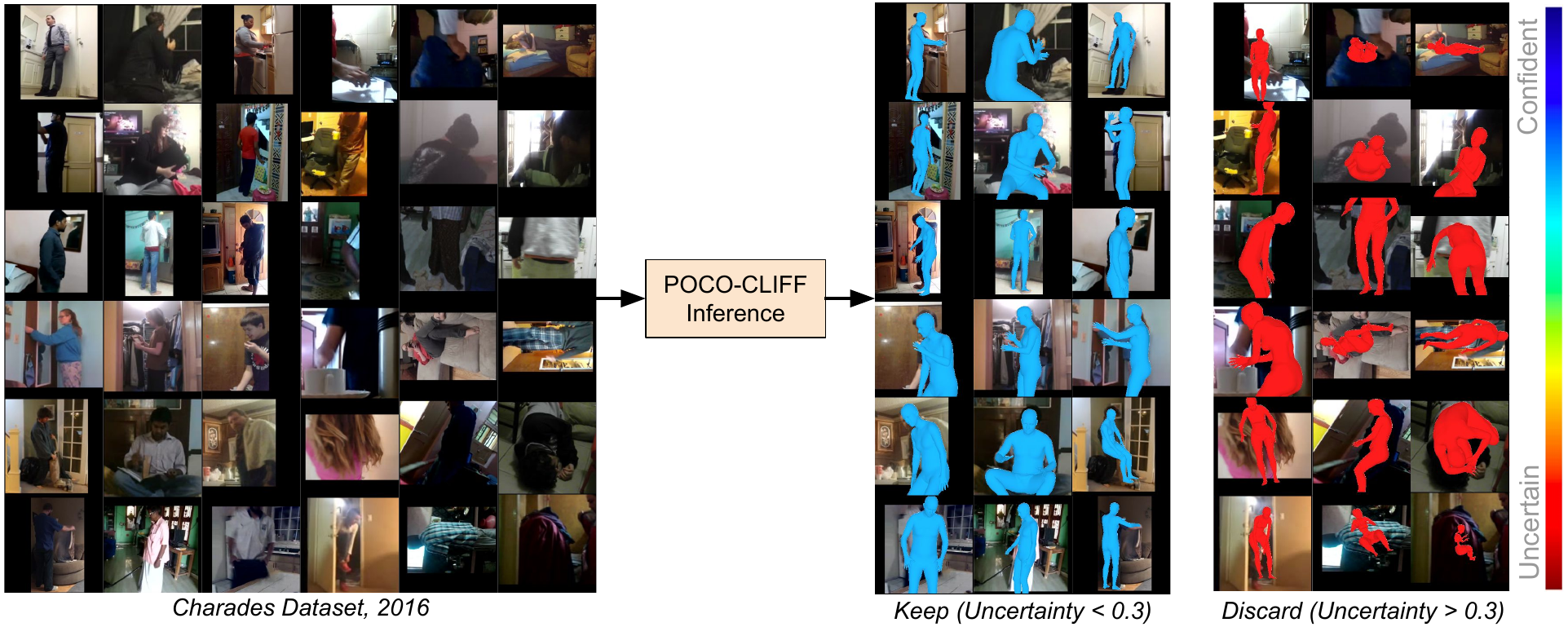}}
    \caption{
        \cheading{Automatic pseudo-GT}
        Random samples of pseudo-GT generated by \modelname-\cliff on Charades \cite{charades} dataset.
        We keep the frames with lower uncertainty and treat the output \smpl parameters as 
        pseudo \groundtruth for re-training.
    }
    \label{fig:supmat_charades_pGT}
\end{figure*}

\begin{figure*}
    \centerline{\includegraphics[width=0.9\textwidth]{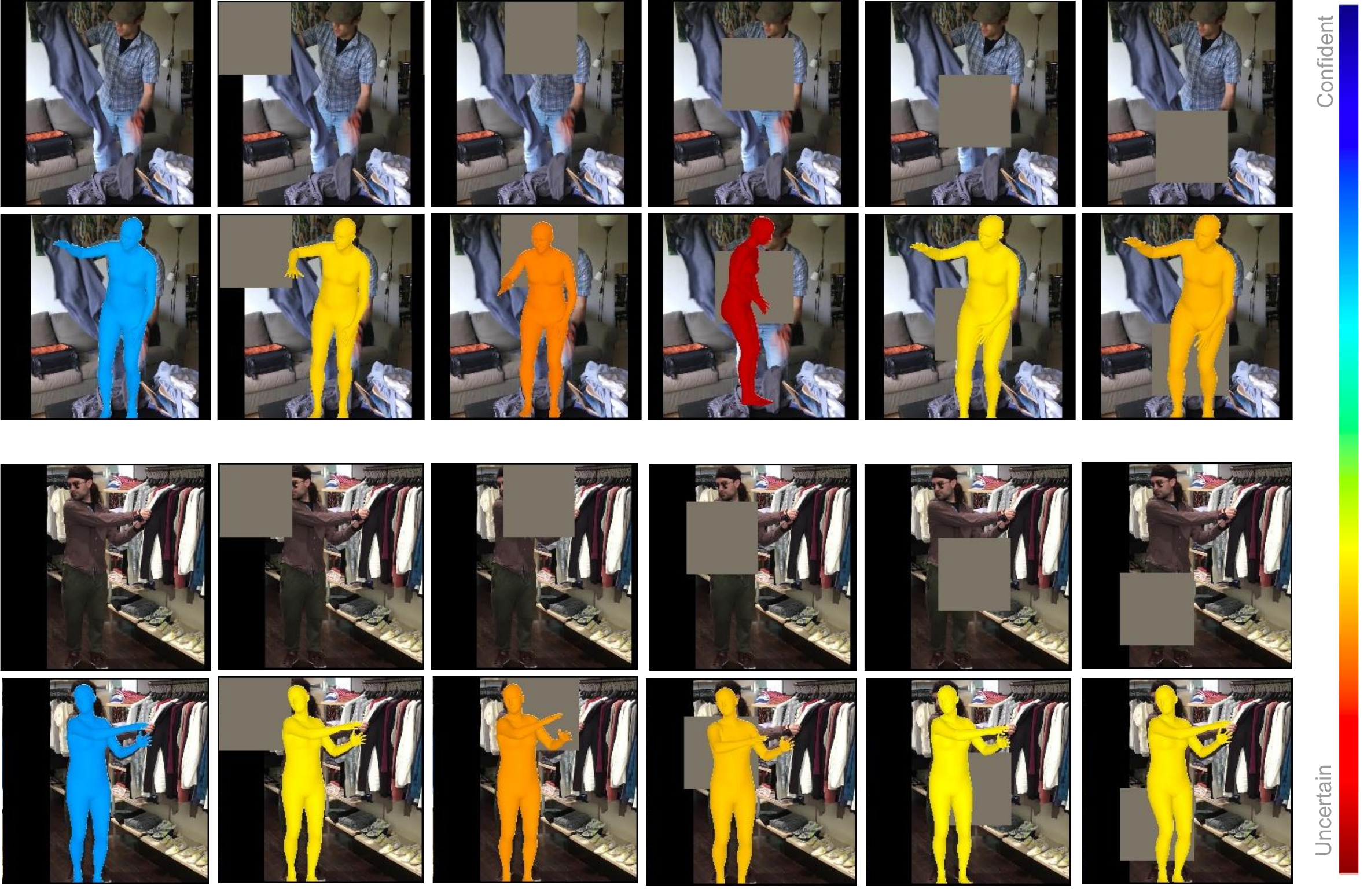}}
    \caption{
        \cheading{Effect of occlusion on uncertainty}
        When an image becomes ambiguous due to a synthetic occluder, \mbox{\modelname-\hmreft} estimates a higher uncertainty. 
    }
    \label{fig:supmat_uncertainty_analysis}
\end{figure*}

\begin{figure*}
    \centerline{\includegraphics[width=0.88\textwidth]{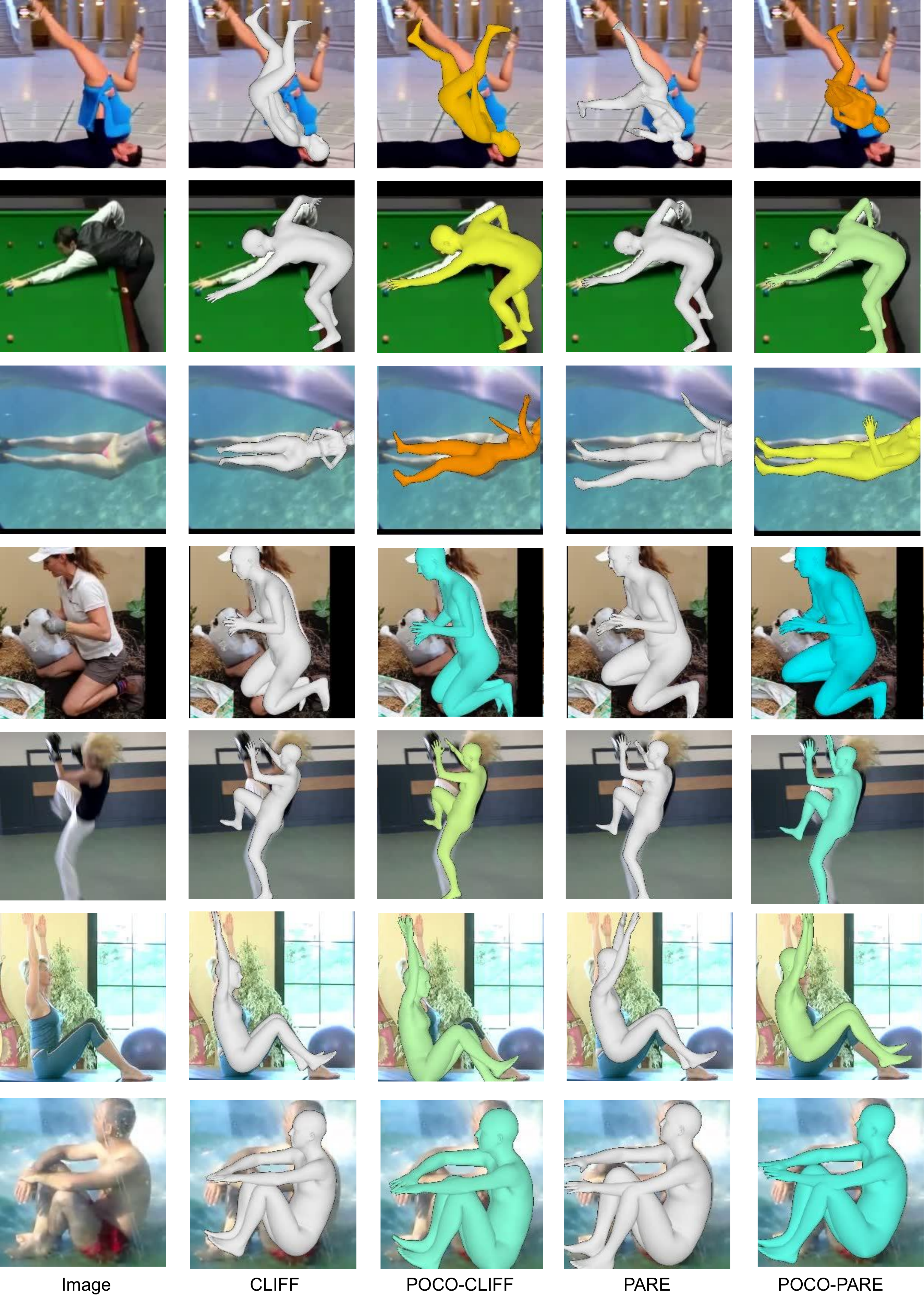}}
    \caption{
        \cheading{Qualitative evaluation for \inthewild images}
        We show results for \cliff~\cite{cliff}, \PARE~\cite{pare} and \modelname versions of respective \hps methods \ie \modelname-\cliff and 
        \modelname-\pare.
    }
    \label{fig:supmat_qual_pare}
\end{figure*}
\begin{figure*}
     \vspace{-0.3 em}
    \centerline{\includegraphics[width=0.88\textwidth]{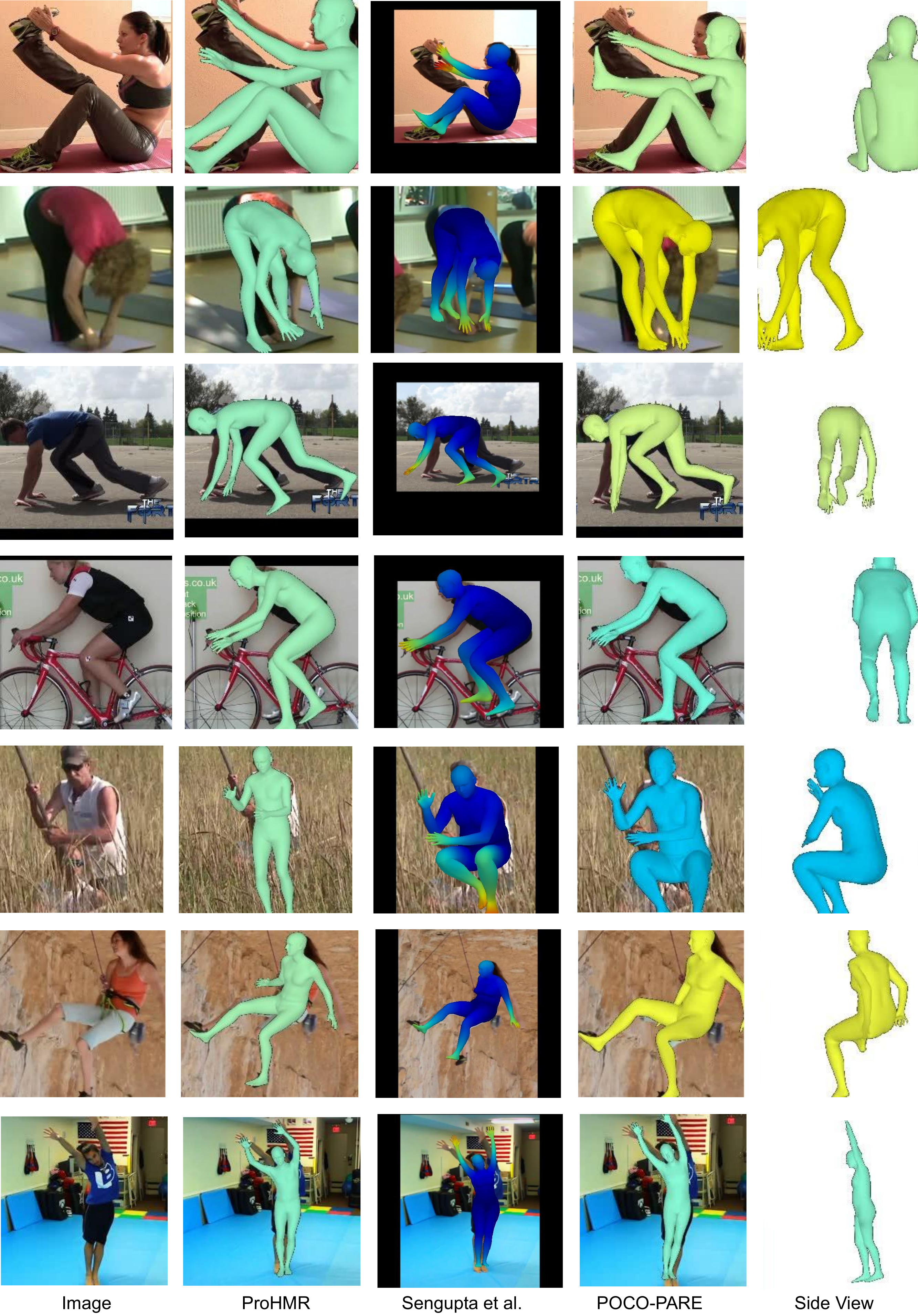}}
    \caption{
        \cheading{Qualitative evaluation for \inthewild images}
        We show results %
        for 
        ProHMR       \cite{prohmr},
        Sengupta \etal \cite{hier_mesh_recovery},
        and 
        our \modelname-\pare. 
    }
    \label{fig:supmat_qual_hph_prohmr}
\end{figure*}

We qualitatively compare 
\modelname with 
the deterministic \HPS methods
like \cliff~\cite{cliff},
\pare~\cite{pare}, and the probabilistic methods ProHMR~\cite{prohmr} and Sengupta \etal~\cite{hier_mesh_recovery}. 
The results are shown in ~\cref{fig:supmat_qual_pare} and ~\cref{fig:supmat_qual_hph_prohmr}, respectively. For more qualitative examples, please see the \video.

\end{document}